%% file: arxiv.tex
\documentclass[11pt]{article}

\pdfoutput=1
\usepackage{geometry}
\usepackage{microtype}
\usepackage{graphicx}
\usepackage{multirow}
\usepackage{subfigure}
\usepackage{booktabs} 
\usepackage{xfrac}
\usepackage{enumitem}
\usepackage{hyperref}

\usepackage{algorithm}
\usepackage{algorithmicx,algpseudocode}

\makeatletter
\def\ALG@special@indent{%
    \ifdim\ALG@thistlm=0pt\relax
        \hskip-\leftmargin
    \else
        \hskip\ALG@thistlm
    \fi
}
\algnewcommand\algorithmicinput{\textbf{Input:}}
\algnewcommand\Input{\item[\algorithmicinput]}

\usepackage[utf8]{inputenc} 
\usepackage[T1]{fontenc}    
\usepackage{url}
\usepackage{pifont}
\usepackage{amsmath}
\usepackage{subfigure}
\usepackage{caption}
\usepackage{hyperref} 

\usepackage{environ}
\newcommand{\acksection}{\section*{Acknowledgments and Disclosure of Funding}}
\NewEnviron{ack}{%
  \acksection
  \BODY
}

\usepackage{mathtools}
\usepackage{xcolor,colortbl}
\def\X{\mathcal{X}}
\def\zd{d_{z}}
\def\dz{d_{z}}
\def\cd{d}
\def\dc{d}
\def\A{\mathcal{A}}
\def\B{\mathcal{B}}
\def\R{\mathbb{R}}
\def\P{\mathbb{P}}
\def\yy{\tilde{y}}
\usepackage{amssymb}

\newenvironment{itemize*}%
{\begin{itemize}[leftmargin=*,topsep=0pt]%
		\setlength{\itemsep}{1pt}%
		\setlength{\parskip}{1pt}}%
	{\end{itemize}}

\usepackage{enumerate}
\usepackage{enumitem}
\def\R{{\mathbb{R}}}
\usepackage{mathtools}
\newcommand{\norm}[1]{\left\lVert#1\right\rVert}

\algnewcommand{\IfThenElse}[3]{ \IfThenElse{<if>}{<then>}{<else>}
  \STATE \algorithmicif\ #1\ \algorithmicthen\ #2\ \algorithmicelse\ #3}

\newcommand{\Stage}[1]{\item[]\noindent\ALG@special@indent \textbf{Initialization:}\ #1}
\usepackage{amsthm}

\usepackage[normalem]{ulem}
\usepackage[export]{adjustbox}

\usepackage{xcolor}
\theoremstyle{plain}
\newtheorem{theorem}{Theorem}[section]

\newtheorem{lemma}[theorem]{Lemma}
\newtheorem{corollary}[theorem]{Corollary}
\theoremstyle{definition}
\newtheorem{definition}[theorem]{Definition}

\theoremstyle{remark}
\newtheorem{remark}[theorem]{Remark}
\usepackage[textsize=tiny]{todonotes}
\usepackage{ulem}

\title{Robust Lipschitz Bandits to Adversarial Corruptions}

\author
{
Yue Kang\thanks{Department of Statistics, University of California, Davis; e-mail: {\tt yuekang@ucdavis.edu}} 
	\and 
 Cho-Jui Hsieh\thanks{Department of Computer Science, University of California, Los Angeles; e-mail: {\tt chohsieh@cs.ucla.edu}} 
	\and
 Thomas C. M. Lee\thanks{Department of Statistics, University of California, Davis; e-mail: {\tt tcmlee@ucdavis.edu}} 
}

\begin{document}

\date{}
\maketitle

\begin{abstract}
Lipschitz bandit is a variant of stochastic bandits that deals with a continuous arm set defined on a metric space, where the reward function is subject to a Lipschitz constraint.
In this paper, we introduce a new problem of Lipschitz bandits in the presence of adversarial corruptions where an adaptive adversary corrupts the stochastic rewards up to a total budget $C$. The budget is measured by the sum of corruption levels across the time horizon $T$. We consider both weak and strong adversaries, where the weak adversary is unaware of the current action before the attack, while the strong one can observe it. Our work presents the first line of robust Lipschitz bandit algorithms that can achieve sub-linear regret under both types of adversary, even when the total budget of corruption $C$ is unrevealed to the agent. We provide a lower bound under each type of adversary, and show that our algorithm is optimal under the strong case. Finally, we conduct experiments to illustrate the effectiveness of our algorithms against two classic kinds of attacks.
\end{abstract}

\input{introduction.tex}

\input{related.tex}
\input{preliminaries.tex}

\input{known.tex}
\input{agnostic.tex}
\input{exp.tex}
\input{conclusion.tex}

\bibliographystyle{plain}
\bibliography{ref}

\clearpage


\appendix

\clearpage
\section{Appendix}
\input{app_thmrobust.tex}
\input{app_thmrmel.tex}
\input{app_thmmodelselection.tex}
\input{app_thmbob.tex}
\input{app_alteralg.tex}
\input{app_lowerbound.tex}
\input{app_exp.tex}

\end{document}

%% file: introduction.tex
\section{Introduction}\label{sec:intro}

Multi-armed Bandit (MAB)~\cite{auer2002finite} is a fundamental and powerful framework in sequential decision-making problems. 
Given the potential existence of malicious users in real-world scenarios~\cite{Chen2022-cv}, 
a recent line of works considers the stochastic bandit problem under adversarial corruptions: an agent adaptively updates its policy to choose an arm from the arm set, and an adversary may contaminate the reward generated from the stochastic bandit before the agent could observe it. To robustify bandit learning algorithms under adversarial corruptions, several algorithms have been developed in the setting of 
traditional MAB~\cite{gupta2019better,jun2018adversarial,lykouris2018stochastic} and contextual linear bandits~\cite{bogunovic2021stochastic,ding2022robust,he2022nearly,li2019stochastic,zhao2021linear}. These works consider either the weak adversary~\cite{lykouris2018stochastic}, which has access to all past data but not the current action before choosing its attack, or the strong adversary~\cite{bogunovic2021stochastic}, which is also aware of the current action for contamination. Details of these two adversaries will be elaborated in Section~\ref{sec:preliminaries}. In practice, bandits under adversarial corruptions can be used in many real-world problems such as pay-per-click advertising with click fraud and recommendation systems with fake reviews~\cite{lykouris2018stochastic}, and it has been empirically validated that stochastic MABs are vulnerable to slight corruption~\cite{ding2022robust,garcelon2020adversarial,jun2018adversarial}.

Although there has been extensive research on the adversarial robustness of stochastic bandits, most existing works consider problems with a discrete arm set, such as the traditional MAB and contextual linear bandit. In this paper, we investigate robust bandit algorithms against adversarial corruptions in the Lipschitz bandit setting, where a continuously infinite arm set lie in a known metric space with covering dimension $d$ and the expected reward function is an unknown Lipschitz function. Lipschitz bandit can be used to efficiently model many real-world tasks such as dynamic pricing, auction bidding~\cite{slivkins2019introduction} and hyperparameter tuning~\cite{kang2023online}.
The stochastic Lipschitz bandit has been well understood after a large body of literature~\cite{bubeck2008online,kleinberg2019bandits,magureanu2014lipschitz}, and state-of-the-art algorithms could achieve a cumulative regret bound of order $\tilde{O}(T^{\frac{\zd+1}{\zd+2}})$\footnote{$\tilde{O}$ ignores the polylogarithmic factors. $\zd$ is the zooming dimension defined in Section \ref{sec:preliminaries}.} in time $T$. However, to the best of our knowledge, the stochastic Lipschitz bandit problem with adversarial corruptions has never been explored, and we believe it is challenging since most of the existing robust MAB algorithms utilized the idea of elimination, which is much more difficult under a continuously infinite arm pool. 
Furthermore, the complex structure of different metric spaces also poses challenges for defending against adversarial attacks~\cite{solomon2021fast} in theory. Therefore, it remains intriguing to design computationally efficient Lipschitz bandits that are robust to adversarial corruptions under both weak and strong adversaries. 

\begin{table}[t]
\caption{Comparisons of regret bounds for our proposed robust Lipschitz bandit algorithms.}
\label{tab:algorithm}
\vskip 0.06in
\begin{center}
\begin{small}
\begin{tabular}{l|cc|c|c}
\toprule
    \textsc{Algorithm} & \textsc{Regret Bound} & \textsc{Format} & $C$  & \textsc{Adversary}  \\
\midrule
    Robust Zooming & $\tilde O \left(T^{\frac{\zd+1}{\zd+2}} + C^{\frac{1}{\zd+1}} T^{\frac{\zd}{\zd+1}}\right)$ & \textsc{High. Prob.} & \textsc{Known}  & \textsc{Strong} \\
\midrule
    RMEL & $ \tilde O\left((C^\frac{1}{\zd+2} + 1)  T^\frac{\zd+1}{\zd+2} \right)$ & \textsc{High. Prob.} & \textsc{Unknown}  & \textsc{Weak} \\
    {EXP3.P} & $\tilde O\left((C^{\frac{1}{\cd+2}}+1) T^{\frac{\cd+2}{\cd+3}}\right)$ & \textsc{Expected} & \textsc{Unknown}  & \textsc{Strong} \\
    CORRAL & $\tilde O\left((C^{\frac{1}{\cd+1}}+1) T^{\frac{\cd+1}{\cd+2}}\right)$  & \textsc{Expected} & \textsc{Unknown} & \textsc{Strong} \\
    BoB Robust Zooming & $\tilde O\left(T^\frac{\cd+3}{\cd+4} + C^\frac{1}{\cd+1} T^{\frac{\cd+2}{d+3}}\right)$ & \textsc{High. Prob.} & \textsc{Unknown} & \textsc{Strong} \\
\bottomrule
\end{tabular}
\end{small}
\end{center}
\vskip -0.12in
\end{table}

We develop efficient robust algorithms whose regret bounds degrade sub-linearly in terms of the corruption budget $C$.
Our contributions can be summarized as follows: (1) Under the weak adversary, we extend the idea in~\cite{lykouris2018stochastic} and propose an efficient algorithm named Robust Multi-layer Elimination Lipschitz bandit algorithm (RMEL) that is agnostic to $C$ and attains $\tilde O (C^{\frac{1}{\zd+2}}T^{\frac{\zd+1}{\zd+2}})$ regret bound. This bound matches the minimax regret bound of Lipschitz bandits~\cite{bubeck2008online,kleinberg2019bandits} in the absence of corruptions up to logarithmic terms. This algorithm consists of multiple parallel sub-layers with different tolerance against the budget $C$, where each layer adaptively discretizes the action space and eliminates some less promising regions based on its corruption tolerance level in each crafted epoch. Interactions between layers assure the promptness of the elimination process. (2) Under the strong adversary, we first show that when the budget $C$ is given, a simple modification on the classic Zooming algorithm~\cite{kleinberg2019bandits} would lead to a robust method, namely, Robust Zooming algorithm, which could obtain a regret bound of order $\tilde O (T^{\frac{\zd+1}{\zd+2}} + C^{\frac{1}{\zd+1}}T^{\frac{\zd}{\zd+1}})$. We then provide a lower bound to prove the extra $O (C^{\frac{1}{\zd+1}}T^{\frac{\zd}{\zd+1}})$ regret is unavoidable. Further, inspired by the Bandit-over-Bandit (BoB) model selection idea~\cite{cheung2019learning,ding2022syndicated,pacchiano2020model}, we design a two-layer framework adapting to the unknown $C$ where a master algorithm in the top layer dynamically tunes the corruption budget for the Robust Zooming algorithm. Three types of master algorithms are discussed and compared in both theory and practice. Table \ref{tab:algorithm} outlines our algorithms as well as their regret bounds under different scenarios.

%% file: related.tex
\section{Related Work}\label{sec:related}

\paragraph{Stochastic and Adversarial Bandit} Extensive studies have been conducted on MAB and its variations, including linear bandit~\cite{abbasi2011improved}, matrix bandit~\cite{kang2022efficient}, etc. The majority of literature can be categorized into two types of models~\cite{lattimore2020bandit}: stochastic bandit, in which rewards for each arm are independently sampled from a fixed distribution, and adversarial bandit, where rewards are maliciously generated at all time. However, adversarial bandit differs from our problem setting in the sense that rewards are arbitrarily chosen without any budget or distribution constraint. Another line of work aims to obtain ``the best of both worlds'' guarantee simultaneously~\cite{bubeck2012best}.
 However, neither of these models is reliable in practice~\cite{cao2019nearly}, since the former one is too ideal, while the latter one remains very pessimistic, assuming a fully unconstrained setting. Therefore, it is more natural to consider the scenario that lies "in between" the two extremes: the stochastic bandit under adversarial corruptions.

\paragraph{Lipschitz Bandit} Most existing works on the stochastic Lipschitz bandit~\cite{agrawal1995continuum} follow two key ideas. One is to uniformly discretize the action space into a mesh in the initial phase so that any MAB algorithm could be implemented~\cite{kleinberg2004nearly,magureanu2014lipschitz}. The other is to adaptively discretize the action space by placing more probes in more promising regions, and then UCB~\cite{bubeck2008online,kleinberg2019bandits,lu2019optimal}, TS~\cite{kang2023online} or elimination~\cite{feng2022lipschitz} method could be utilized to deal with the exploration-exploitation tradeoff. The adversarial Lipschitz bandit was recently introduced and solved in~\cite{podimata2021adaptive}, where the expected reward Lipschitz function is arbitrarily chosen at each round. However, as mentioned in the previous paragraph, this fully adversarial setting is quite different from ours. And their algorithm relies on several unspecified hyperparameters and hence is computationally formidable in practice.
 In addition, some interesting variations of Lipschitz bandits have also been carefully examined, such as contextual Lipschitz bandits~\cite{slivkins2011contextual}, full-feedback Lipschitz bandits~\cite{kleinberg2010sharp}, and taxonomy bandits~\cite{slivkins2011multi}.

\paragraph{Robust Bandit to Adversarial Corruptions} Adversarial attacks were studied in the setting of MAB~\cite{jun2018adversarial} and linear bandits~\cite{garcelon2020adversarial}. And we will use two classic attacks for experiments in Section~\ref{sec:exp}. To defend against attacks from weak adversaries, \cite{lykouris2018stochastic} proposed the first MAB algorithm robust to corruptions with a regret $C$ times worse than regret in the stochastic setting. An improved algorithm whose regret only contains an additive term on $C$ was then proposed in \cite{gupta2019better}. \cite{li2019stochastic} subsequently studied the linear bandits with adversarial corruptions and achieved instance-dependent regret bounds. \cite{lee2021achieving} also studied the corrupted linear bandit problem while assuming the attacks on reward are linear in action. Recently, a robust VOFUL algorithm achieving regret bound only logarithmically dependent on $T$ was proposed in \cite{wei2022model}. Another line of work on the robust bandit problem focuses on a more challenging setting with strong adversaries who could observe current actions before attacking rewards. \cite{bogunovic2021stochastic} considered the corrupted linear bandit when small random perturbations are applied to context vectors, and \cite{ding2022robust,he2022nearly,zhao2021linear} extended the OFUL algorithm~\cite{abbasi2011improved} and achieved improved regret bounds. \cite{ye2023corruption} further considers general non-linear contextual bandits and also MDPs with strong adversarial corruptions. However, the study of Lipschitz bandits under attacks remains an unaddressed open area.

%% file: preliminaries.tex
\section{Preliminaries}\label{sec:preliminaries}
We will introduce the setting of Lipschitz bandits with adversarial corruptions in this section. The Lipschitz bandit is defined on a triplet $(\X,D,\mu)$, where $\X$ is the arm set space equipped with some metric $D$, and $\mu: \X \rightarrow \R$ is an unknown Lipschitz reward function on the metric space $(\X,D)$ with Lipschitz constant $1$. W.l.o.g. we assume $\X$ is compact with its diameter no more than $1$. Under the stochastic setting, at each round $t \in [T] \coloneqq \{1,2,\dots,T\}$, stochastic rewards are sampled for each arm $x \in \X$ from some unknown distribution $\P_x$ independently, and then the agent pulls an arm $x_t$ and receives the corresponding stochastic reward $\tilde{y}_t$ such that,
\begin{align}
    \tilde{y}_t = \mu(x_t) + \eta_t, \label{eq:problem}
\end{align}
where $\eta_t$ is i.i.d. zero-mean random error with sub-Gaussian parameter $\sigma$ conditional on the filtration $\mathcal{F}_t = \{x_t,x_{t-1},\eta_{t-1},\dots,x_1,\eta_1\}$. W.l.o.g we assume $\sigma = 1$ for simplicity in the rest of our analysis. At each round $t\in[T]$, the weak adversary observes the payoff function $\mu(\cdot)$, the realizations of $\P_x$ for each arm $x\in\X$ and choices of the agent $\{x_i\}_{i=1}^{t-1}$ in previous rounds, and injects an attack $c_t(x)$ into the reward before the agent pulls $x_t$. The agent then receives a corrupted reward $y_t = \yy_t+c_t(x_t)$. The strong adversary would be omniscient and have complete information about the problem $\mathcal{F}_t$. In addition to the knowledge that a weak adversary possesses, it would also be aware of the current action $x_t$ while contaminating the data, and subsequently decide upon the corrupted reward $y_t = \yy_t+c_t(x_t)$. Some literature in corrupted bandits~\cite{ding2022robust,garcelon2020adversarial} also consider attacking on the contexts or arms, i.e. the adversary modifies the true arm $x_t$ in a small region, while in our problem setting it is obvious that attacking contexts is only a sub-case of attacking rewards due to the Lipschitzness of $\mu(\cdot)$, and hence studying the adversarial attacks on rewards alone is sufficient under the Lipschitz bandit setting.

The total corruption budget $C$ of the adversary is defined as $C = \sum_{t=1}^T \max_{x\in\X} |c_t(x)|$, which is the sum of maximum perturbation from the adversary at each round across the horizon $T$. Note the strong adversary may only corrupt the rewards of pulled arms and hence we could equivalently write $C = \sum_{t=1}^T |c_t(x_t)|$ in that case as~\cite{bogunovic2021stochastic,he2022nearly}.
Define the optimal arm $x_* = \arg\max_{x\in\X} \mu(x)$ and the loss of arm $x$ as $\Delta(x) = \mu(x_*) \! - \! \mu(x), x \in\X$. W.l.o.g. we assume $C \leq T$ and each instance of attack $|c_t(x)|\leq 1, \forall t \in [T], x \in \X$ as in other robust bandit literature~\cite{gupta2019better,lykouris2018stochastic} since the adversary could already make any arm $x \in \X$ optimal given that $\Delta(x) \leq 1$. (We can assume $|c_t(x)|\leq u, \forall t \in [T], x \in \X$ for any positive constant $u$.) 
Similar to the stochastic case~\cite{kleinberg2004nearly}, the goal of the agent is to minimize the cumulative regret defined as:
\begin{align}
    Regret_T = T \mu(x_*) - \sum_{t=1}^T \mu(x_t). \label{eq:regret}
\end{align}
An important pair of concepts in Lipschitz bandits defined on $(\X,D,\mu)$ are the covering dimension $\cd$ and the zooming dimension $\zd$. Let $\B(x,r)$ denotes a closed ball centered at $x$ with radius $r$ in $\X$, i.e. $\B(x,r) = \{x'\in\X:D(x,x') \leq r\}$, the $r$-covering number $N_c(r)$ of metric space $(\X,D)$ is defined as the minimal number of balls with radius of no more than $r$ required to cover $X$. On the contrary, the $r$-zooming number $N_z(r)$ introduced in~\cite{kleinberg2019bandits} not only depends on the metric space $(\X,D)$ but also the payoff function $\mu(\cdot)$. It describes the minimal number of balls of radius not more than $r/16$ required to cover the $r$-optimal region defined as $\{x\in\X:\Delta(x) \leq r \}$~\cite{bubeck2008online,feng2022lipschitz}\footnote{We actually use the near-optimality dimension introduced in~\cite{bubeck2008online}, where the authors imply the equivalence between this definition and the original zooming dimension proposed in~\cite{kleinberg2019bandits}.}. Next, we define the covering dimension $\cd$ (zooming dimension $\zd$) as the smallest $q \geq 0$ such that for every $r\in(0,1]$ the $r$-covering number $N_c(r)$ ($r$-zooming number $N_z(r)$) can be upper bounded by $\alpha r^{-q}$ for some multiplier $\alpha>0$ that is free of $r$:
\begin{gather*}
\cd = \min \{q \geq 0: \exists \alpha>0, N_c(r) \leq \alpha r^{-q}, \forall r \in (0,1] \}, \\
\zd = \min \{q \geq 0: \exists \alpha>0, N_z(r) \leq \alpha r^{-q}, \forall r \in (0,1] \}.
\end{gather*}
It is clear that $0\leq \zd \leq \cd$ since the $r$-optimal region is a subset of $\X$. On the other hand, $\zd$ could be much smaller than $\cd$ in some benign cases. For example, if the payoff function $\mu(\cdot)$ defined on the metric space $(\R^k,\norm{\cdot}_2), k \in \mathbb{N}$ is $C^2$-smooth and strongly concave in a neighborhood of the optimal arm $x_*$, then it could be easily verified that $\zd=k/2$ whereas $\dc = k$. However, $\dz$ is never revealed to the agent as it relies on the underlying function $\mu(\cdot)$, and hence designing an algorithm whose regret bound depends on $\zd$ without knowledge of $\zd$ would be considerably difficult.

%% file: known.tex
\section{Warm-up: Robust Lipschitz Bandit with Known Budgets}\label{sec:known}
\begin{algorithm}[t]
\caption{Robust Zooming Algorithm} \label{alg:robust}
\begin{algorithmic}[1]
\Input Arm metric space $(\X,D)$, time horizon $T$, probability rate $\delta$.
\Stage  {Active arm set $J = \{\}$, active space $\X_{act} = \X$.}
\For{$t=1$ {\bfseries to} $T$}
\If{$f(v)-f(u) \geq r(v)+2r(u)$ for some pair of active arms $u,v \in J$.}
\State{Set $J = J \, \backslash \{u\}$ and $\X_{act} = \X_{act} \, \backslash \B(u,r(u))$.} {\color{red}\Comment{Removal}}
\EndIf
\If{$\X_{act} \nsubseteq \cup_{v \in J} \B(v,r(v))$}
\State {Activate and pull some arm $x \notin \cup_{v \in J} \B(v,r(v))$ in $\X_{act}$ such that $x_t = x, \; J = J \cup \{x\}$, and set the components $n(x)=0,\,f(x)=0$.}   {\color{red}\Comment{Activation}}
\Else
\State{Pull $x_t = \arg\max_{v \in J} I(v)=f(v)+2r(v)$, and break ties arbitrarily.} {\color{red}\Comment{Selection}}
\EndIf
\State Observe the payoff $y_t$. And update components associated with $x_t$ in the Robust Zooming Algorithm: $n(x_t) = n(x_t)+1, f(x_t) = \left(f(x_t)\left(n(x_t)-1\right)+y_t \right)/n(x_t)$. 
\EndFor
\end{algorithmic}
\end{algorithm}
To defend against attacks on Lipschitz bandits, we first consider a simpler case where the agent is aware of the corruption budget $C$. We demonstrate that a slight modification of the classic Zooming algorithm~\cite{kleinberg2019bandits} can result in a robust Lipschitz bandit algorithm even under the strong adversary, called the Robust Zooming algorithm, which achieves a regret bound of order $\tilde O (T^{\frac{\zd+1}{\zd+2}} + C^{\frac{1}{\zd+1}} T^{\frac{\zd}{\zd+1}})$. 

We first introduce some notations of the algorithm: denote $J$ as the active arm set. For each active arm $x\in J$, let $n(x)$ be the number of times arm $x$ has been pulled, $f(x)$ be the corresponding average sample reward, and $r(x)$ be the confidence radius controlling the deviation of the sample average $f(x)$ from its expectation $\mu(x)$. We also define $\B(x,r(x))$   as the confidence ball of an active arm $x$. In essence, 
the Zooming algorithm works by focusing on regions that have the potential for higher rewards and allocating fewer probes to less promising regions.
 The algorithm consists of two phases: in the activation phase, a new arm gets activated if it is not covered by the confidence balls of all active arms. This allows the algorithm to quickly zoom into the regions where arms are frequently pulled due to their encouraging rewards. In the selection phase, the algorithm chooses an arm with the largest value of $f(v)+2r(v)$ among $J$ based on the UCB methodology.

Our key idea is to enlarge the confidence radius of active arms to account for the known corruption budget $C$. Specifically, we could set the value of $r(x)$ as:
$$r(x) = \sqrt{\frac{4 \ln{(T)} + 2 \ln{(2/\delta)}}{n(x)}} + \frac{C}{n(x)},$$
where the first term accounts for deviation in stochastic rewards and the second term is used to defend the corruptions from the adversary. The robust algorithm is shown in Algorithm \ref{alg:robust}. In addition to the two phases presented above, our algorithm also conducts a removal procedure at the beginning of each round for better efficiency. This step adaptively  removes regions that are likely to yield low rewards with high confidence. Theorem \ref{thm:robust} provides a regret bound for Algorithm \ref{alg:robust}.

\begin{theorem}\label{thm:robust}
Given the total corruption budget that is at most $C$,  with probability at least $1-\delta$, the overall regret of Robust Zooming Algorithm (Algorithm \ref{tab:algorithm}) can be bounded as:
$$Regret_T =  O\left( \ln{(T)}^{\frac{1}{\zd+2}} T^{\frac{\zd+1}{\zd+2}} + C^{\frac{1}{\zd+1}} T^{\frac{\zd}{\zd+1}} \right) = \tilde O\left( T^{\frac{\zd+1}{\zd+2}} + C^{\frac{1}{\zd+1}} T^{\frac{\zd}{\zd+1}} \right).$$
\end{theorem}
\vspace{-1mm}
Furthermore, the following Theorem~\ref{thm:lowerbounds} implies that our regret bound attains the lower bound and hence is unimprovable. The detailed proof is given in Appendix~\ref{app:lowerbounds}. 

\begin{theorem}\label{thm:lowerbounds}
Under the strong adversary with a corruption budget of $C$, for any zooming dimension $\dz \in \mathbb{Z}^+$, there exists an instance for which any algorithm (even one that is aware of $C$) must suffer a regret of order $\Omega (C^{\frac{1}{\zd+1}}T^{\frac{\zd}{\zd+1}} )$ with probability at least $0.5$.
\end{theorem}

In addition to the lower bound provided in Theorem~\ref{thm:lowerbounds}, we further propose another lower bound for the strong adversary particularly in the case that $C$ is unknown in the following Theorem~\ref{thm:lowerboundsknown}:
\begin{theorem}\label{thm:lowerboundsknown}
For any algorithm, when there is no corruption, we denote $R_T^0$ as the upper bound of cumulative regret in $T$ rounds under our problem setting described in Section~\ref{sec:preliminaries}, i.e. $Regret_T \leq R_T^0$ with high probability, and it holds that $R_T^0 = o(T)$. Then under the strong adversary and unknown attacking budget $C$, there exists a problem instance on which this algorithm will incur linear regret $\Omega(T)$ with probability at least $0.5$, if $C = \Omega(R_T^0/4^{d_z})=\Omega(R_T^0)$.
\end{theorem}


However, there are also some weaknesses to our Algorithm \ref{alg:robust}. The first weakness is that the algorithm is too conservative and pessimistic in practice since the second term of $r(x)$ would dominate under a large given value of $C$. We could set the second term of $r(x)$ as $\min\{1,C/n(x)\}$ to address this issue and the analysis of Theorem \ref{thm:robust} will still hold as shown in Appendix~\ref{app:thmrobust} Remark~\ref{rem:thmrobust}. The second weakness is that it still incurs the same regret bound shown in Theorem \ref{thm:robust} even if there are actually no corruptions applied. To overcome these problems and to further adapt to the unknown corruption budget $C$, we propose two types of robust algorithms in the following Section \ref{sec:agnostic}.

%% file: agnostic.tex
\section{Robust Lipschitz Bandit with Unknown Budgets }\label{sec:agnostic}
In practice, a decent upper bound of the corruption budget $C$ would never be revealed to the agent, and hence it is important to develop robust Lipschitz bandit algorithms agnostic to the amount of $C$. In this section, we first propose an efficient adaptive-elimination-based approach to deal with the weak adversary, and then we adapt our Algorithm \ref{alg:robust} by using several advances of model selection in bandits to defend against the attacks from the strong adversary. 
\subsection{Algorithm for Weak Adversaries}
\label{sec:weak_adversaries}
The weak adversary is unaware of the agent's current action before contaminating the stochastic rewards. We introduce an efficient algorithm called Robust Multi-layer Elimination Lipschitz bandit algorithm (RMEL) that is summarized in Algorithm \ref{alg:rmel}. Four core steps are introduced as follows.

\textbf{Multi-layer Parallel Running:} Inspired by the multi-layer idea used in robust MABs~\cite{lykouris2018stochastic}, our algorithm consists of multiple sub-layers running in parallel, each with a different tolerance level against corruptions. As shown in Algorithm \ref{alg:rmel}, there are $l^*$ layers and the tolerance level of each layer, denoted as $v_l$, increases geometrically with a ratio of $B$ (a hyperparameter). At each round, a layer $l$ is sampled with probability $1/v_l$, meaning that layers that are more resilient to attacks are less likely to be chosen and thus may make slower progress. This sampling scheme helps mitigate  adversarial perturbations across layers by limiting the amount of corruptions distributed to layers whose tolerance levels exceed the unknown budget $C$ to at most $O(\ln(T/\delta))$. 
For the other low-tolerance layers which may suffer from  high volume of attacks, we use the techniques introduced below to rectify them in the guidance of the elimination procedure on robust layers. While we build on the multi-layer idea introduced in~\cite{lykouris2018stochastic}, our work introduces significant refinements and novelty by extending this approach to continuous and infinitely large arm sets, as demonstrated below. 

\textbf{Active Region Mechanism:} For each layer $\ell$, our algorithm proceeds in epochs: 
we initialize the epoch index $m_l = 1$  and construct a $1/2^{m_l}$-covering of $\X$ as the active region set $\A_l$. In addition, we denote $n_l$ as the number of times that layer $l$ has been chosen, and for each active region $A \in \A_l$ we define $n_{l,A}, f_{l,A}$ as the number of times $A$ has been chosen as well as its corresponding average empirical reward respectively. Assume layer $l_t$ is selected at time $t$, then only one active region (denoted as $A_t$) in $\A_{l_t}$ would be played where we arbitrarily pull an arm $x_t \in A_t$ and collect the stochastic payoff $y_t$. For any layer $l$, if each active region in $\A_l$ is played for $6 \ln{(4T/\delta)} \cdot 4^{m}$ times (i.e. line 6 of Algorithm \ref{alg:rmel}), it will progress to the next epoch after an elimination process that is described below. All components mentioned above that are associated with the layer $l$ will subsequently be refreshed (i.e. line 10 of Algorithm \ref{alg:rmel}).

Although the elimination idea has been used for various bandit algorithms~\cite{slivkins2019introduction}, applying it to our problem setting poses significant challenges due to three sources of error: (1) uncertainty in stochastic rewards; (2) unknown corruptions assigned to each layer; (3) approximation bias between the pulled arm and its neighbors in the same active region. We encapsulate our carefully designed elimination procedure that is specified in line 8 of Algorithm \ref{alg:rmel} from two aspects:

\begin{algorithm}[t]
\caption{Robust Multi-layer Elimination Lipschitz Bandit Algorithm (RMEL)} \label{alg:rmel}
\begin{algorithmic}[1]
\Input Arm metric space $(\X,D)$, time horizon $T$, probability rate $\delta$, base parameter $B$.
\Stage Tolerance level $v_l = \ln{(4T/\delta)} B^{l-1}, m_l=1, n_l = 0, \A_l = 1/2$-covering of $\X, \, f_{l,A} = n_{l,A} = 0$ for all $A \in \A_l, l \in [l^*]$ where $l^* \coloneqq \min \{l \in \mathbb{N} : \ln{(4T/\delta)} B^{l-1} \geq T \}$.
\For{$t=1$ {\bfseries to} $T$}
\State Sample layer $l \in [l^*]$ with probability $1/v_l$, with the remaining probability sampling $l=1$. Find the minimum layer index $l_t \geq l$ such that $\A_{l_t} \neq \emptyset$. {\color{red} \Comment{Layer sampling}}
\State Choose $A_t = \arg\min_{A \in \A_{l_t}} n_{l_t,A}$, break ties arbitrary.
\State Randomly pull an arm $x_t \in A_t$, and observe the payoff $y_t$.
\State Set $n_{l_t} = n_{l_t}+1, \, n_{l_t,A_t}= n_{l_t,A_t}+1,\text{and } f_{l_t,A_t} =\left(f_{l_t,A_t} (n_{l_t,A_t}-1) + y_t\right)/n_{l_t,A_t}$.
\If{$n_{l_t} = 6\ln(4T/\delta)\cdot 4^{m_l} \times |\A_{l_t}|$} 
\State Obtain $f_{l_t,*} = \max_{A \in \A_{l_t}} f_{l_t,A}$.
\State For each $A \in \A_{l_t}$, if $f_{l_t,*} -  f_{l_t,A} > 4/2^{m_{l_t}}$, then we eliminate $A$ from $\A_{l_t}$ and all active regions $A'$ from $\A_{l'}$ in the case that $A' \subseteq A, A' \in \A_{l'}, l' < l$. {\color{red}\Comment{Removal}}
\State Find $1/2^{m_l+1}$-covering of each remaining $A \in \A_{l_t}$ in the same way as $A$ was partitioned in other layers. Then reload the active region set $\A_{l_t}$ as the collection of these coverings.
\State Set $n_{l_t} = 0, \, m_{l_t} = m_{l_t}+1$. And renew $n_{l_t,A} = f_{l_t,A} = 0, \forall A \in \A_{l_t}$. {\color{red}\Comment{Refresh}}
\EndIf
\EndFor
\end{algorithmic}
\end{algorithm}

\textbf{Within-layer Region Elimination and Discretization:} For any layer $l \in [l^*]$, the within-layer elimination occurs at the end of each epoch as stated above. We obtain the average empirical reward $f_{l,A}$ for all $A \in \A_l$ and then discard regions with unpromising payoffs compared with the optimal one with the maximum estimated reward (i.e. $f_{l,*}$ defined in line 7 of Algorithm \ref{alg:rmel}). We further ``zoom in'' on the remaining regions of the layer $l$ that yield satisfactory rewards: we divide them into $1/2^{m_l+1}$-covering and then reload $A_l$ as the collection of these new partitions for the next epoch (line 9 of Algorithm~\ref{alg:rmel}) for the layer $l$. 
In consequence, only regions with nearly optimal rewards would remain and be adaptively discretized in the long run. 

\textbf{Cross-layer Region Elimination:} While layers are running in parallel, it is essential to facilitate communication among them to prevent less reliable layers from getting trapped in suboptimal regions. In our Algorithm \ref{alg:rmel}, if an active region $A \in \A_l$ is eliminated based on the aforementioned rule, then $A$ will also be discarded in all layers $l' \leq l$. This is because the lower layers are faster whereas more vulnerable and less resilient to malicious attacks, and hence they should learn from the upper trustworthy layers whose tolerance levels surpass $C$ by imitating their elimination decisions.

Note that the elimination step (line 8 of Algorithm \ref{alg:rmel}) could be executed either after each epoch or after each round. The former, which is used in the current Algorithm \ref{alg:rmel}, is computationally simpler as it does not require the entire elimination procedure to be repeated each time. On the other hand, the latter version is more precise as it can identify and eliminate sub-optimal regions more quickly. We defer the pseudocode and more details of the latter version in Appendix \ref{app:alteralg} due to the space limit. Another tradeoff lies in the selection of the hyperparameter $B$, which controls the ratio of tolerance levels between adjacent layers. With a larger value of $B$, only fewer layers are required, and hence more samples could be assigned to each layer for better efficiency. But the cumulative regret bound would deteriorate since it's associated with $B$ sub-linearly. The cumulative regret bound is presented in the following Theorem \ref{thm:rmel}, with its detailed proof in Appendix~\ref{app:thmrmel}.

\begin{theorem}\label{thm:rmel}
If the underlying corruption budget is $C$, then with probability at least $1-\delta$, the overall regret of our RMEL algorithm (Algorithm \ref{alg:rmel}) could be bounded as:
$$Regret_T = \tilde O\left( \left((BC)^{\frac{1}{\dz+2}} + 1\right) T^{\frac{\zd+1}{\zd + 2}} \right) = \tilde O\left( \left(C^{\frac{1}{\dz+2}} + 1\right) T^{\frac{\zd+1}{\zd + 2}} \right).$$
\end{theorem}
\textit{Proof sketch:} It is highly non-trivial to deduce the regret bound depending on $C$ and $\zd$ in Theorem~\ref{thm:rmel} without the knowledge of these two values. We first bound the cumulative regret occurred in the robust layers whose tolerance levels $v_l$ are no less than the unknown $C$ by showing the estimated mean $f_{l,A}$ is close to the underlying ground truth $\mu(x), x\in A$ for all time $t$ and active regions $A$ simultaneously with high probability, i.e. we define the set $\Phi$ as follows and prove $P(\Phi) \geq 1-3\delta /4$.
\begin{align*}
    \Phi = \Big\{\left\lvert f_{l,A} - \mu(x) \right\rvert \leq\frac{1}{2^{m_l}} + & \sqrt{\frac{4 \ln{(T)} + 2 \ln{(4/\delta)}}{n_{l,A,t}}} + \frac{\ln{(T)} + \ln{(4/\delta)}}{n_{l,A,t}} : \nonumber
    \\& \qquad \qquad \qquad \forall x \in A, \, \forall A \in \A_{l}, \, \forall l \text{ s.t. } v_l \geq C, \, \forall t \in [T] \Big\}.
\end{align*}
This concentration result can guarantee that regions with unfavorable rewards will be adaptively eliminated. And we could show that the extra cumulative regret from the other vulnerable layers be controlled by the regret occurred in the robust layers. A detailed proof is presented in Appendix~\ref{app:thmrmel}.

Note that if no corruption is actually applied (i.e. $C=0$), our RMEL algorithm could attain a regret bound of order $\tilde O(T^{\frac{\zd+1}{\zd + 2}})$ which coincides with the lower bound of stochastic Lipschitz bandits up to logarithmic terms. We further prove a regret lower bound of order $\Omega(C)$ under the weak adversary in Theorem~\ref{thm:lowerboundw} with its detailed proof in Appendix~\ref{app:lowerboundw}. 
Therefore, a compelling open problem is to narrow the regret gap by proposing an algorithm whose regret bound depends on $C$ in another additive term free of $T$ under the weak adversary, like~\cite{gupta2019better} for MABs and~\cite{he2022nearly} for linear bandits.

\begin{theorem}\label{thm:lowerboundw}
Under the weak adversary with corruption budget $C$, for any zooming dimension $\dz$, there exists an instance such that any algorithm (even is aware of $C$) must suffer from the regret of order $\Omega (C)$ with probability at least $0.5$.
\end{theorem}



\subsection{Algorithm for Strong Adversaries}
\vspace{-0.02 cm}

In Section~\ref{sec:known}, we propose the Robust Zooming algorithm to handle the strong adversary given the knowledge of budget $C$ and prove that it achieves the optimal regret bound. However, compared with the known budget $C$ case, defending against strong adversaries naturally becomes more challenging when the agent is unaware of the budget $C$. Motivated by the literature on model selection in bandits, we extend our Robust Zooming algorithm by combining it with different master algorithms to learn and adapt to the unknown $C$ on the fly. We consider two approaches along this line: the first approach uses the master algorithms EXP3.P and CORRAL with the smoothing transformation~\cite{pacchiano2020model} to deal with unknown $C$, which leads to a promising regret bound but a high computational cost. We then equip Robust Zooming algorithm with the efficient bandit-over-bandit (BoB) idea~\cite{cheung2019learning} to adapt to the unknown $C$, leading to a more efficient algorithm with a slightly worse regret bound. 

\textbf{Model Selection:} When an upper bound on $C$ is known, we propose the Robust Zooming algorithm with regret bound $\tilde O (T^{\frac{\zd+1}{\zd+2}} + C^{\frac{1}{\zd+1}} T^{\frac{\zd}{\zd+1}})$ against strong adversaries in Section~\ref{sec:known}. Therefore, it is natural to consider a decent master algorithm that selects between $\lceil \log_2(T) \rceil$ base algorithms where the $i$-th base algorithm is the Robust Zooming algorithm with corruptions at most $2^i$.
As $C \leq T$, there must exist a base algorithm that is at most $2C$-corrupted. Here we choose the stochastic EXP3.P and CORRAL with smooth transformation proposed in~\cite{pacchiano2020model} as the master algorithm due to the following two reasons with respect to theoretical analysis: (1). our action set $\A$ is fixed and the expected payoff is a function of the chosen arm, which satisfies the restrictive assumptions of this master algorithm (Section 2,~\cite{pacchiano2020model}); (2). the analysis in \cite{pacchiano2020model} still works even the regret bounds of base algorithms contain unknown values, and note the regret bound of our Zooming Robust algorithm depends on the unknown $C$. Based on Theorem 3.2 in~\cite{pacchiano2020model}, the expected cumulative regret of our Robust Zooming algorithm with these two types of master algorithms could be bounded as follows:
\begin{theorem} \label{thm:model_selection}
     When the corruption budget $C$ is unknown, by using our Algorithm \ref{alg:robust} with $\{2^i\}_{i=1}^{\lceil \log_2(T) \rceil}$ corruptions as base algorithms and the EXP3.P and CORRAL with smooth transformation~\cite{pacchiano2020model} as the master algorithm, the expected regret could be upper bounded by
$$\mathbb{E}(Regret_T) = \begin{cases}
\tilde O\left((C^{\frac{1}{\cd+2}}+1) T^{\frac{\cd+2}{\cd+3}}\right) \quad \; \; \text{EXP3.P},\\
\tilde O\left((C^{\frac{1}{\cd+1}}+1) T^{\frac{\cd+1}{\cd+2}}\right)  \quad \; \; \text{CORRAL}.
\end{cases}$$
\end{theorem}
We can observe that the regret bounds given in Theorem~\ref{thm:model_selection} are consistent with the lower bounds presented in Theorem~\ref{thm:lowerboundsknown}. And CORRAL is better under small corruption budgets $C$ (i.e. $C = \tilde O(T^{\frac{\cd+1}{\cd+3}})$) whereas EXP3.P is superior otherwise. Note that the order of regret relies on $\cd$ instead of $\zd$ since the unknown $\zd$ couldn't be used as a parameter in practice, and both regret bounds are worse than the lower bound given in Theorem~\ref{thm:lowerbounds} for the strong adversary. Another drawback of the above method is that a two-step smoothing procedure is required at each round, which is computationally expensive. Therefore, for better practical efficiency, we propose a simple BoB-based method as follows:

\textbf{BoB Robust Zooming:} The BoB idea~\cite{cheung2019learning} is a special case of model selection in bandits and aims to adjust some unspecified hyperparameters dynamically in batches. Here we use $\lceil \log_2(T) \rceil$ Robust Zooming algorithms with different corruption levels shown above as base algorithms in the bottom layer and the classic EXP3.P~\cite{auer2002nonstochastic} as the top layer. Our method, named BoB Robust Zooming, divides $T$ into $H$ batches of the same length, and in one batch keeps using the same base algorithm that is selected from the top layer at the beginning of this batch. When a batch ends, we refresh the base algorithm and use the normalized accumulated rewards of this batch to update the top layer EXP3.P since the EXP3.P algorithm~\cite{auer2002nonstochastic} requires the magnitude of rewards should at most be $1$ in default. Specifically, we normalize the cumulative reward at the end of each batch by dividing it with $(2H + \sqrt{2H \log({12T}/{H \delta})})$ due to the fact that the magnitude of the cumulative reward at each batch would at most be this value with high probability as shown in Lemma~\ref{lem:bound} in Appendix~\ref{app:thmbob}.
Note that this method is highly efficient since a single update of the EXP3.P algorithm only requires $O(1)$ time complexity, and hence the additional computation from updating EXP3.P is only $O(H)$. Due to space limit, we defer Algorithm~\ref{alg:bob} to Appendix \ref{app:alteralg}, and the regret bound is given as follows:
\begin{theorem} \label{thm:bob}
 When the corruption budget $C$ is unknown, with probability at least $1-\delta$, the regret of our BoB Robust Zooming algorithm with $H = T^{(\cd+2)/(\cd+4)}$ could be bounded as:
$$Regret_T = \tilde O\left(T^{\frac{\cd+3}{\cd+4}} + C^{\frac{1}{\cd+1}} T^{\frac{\cd+2}{\cd+3}}\right).$$
\end{theorem}
Although we could deduce the more challenging high-probability regret bound for this algorithm, its order is strictly worse than those given in Theorem~\ref{thm:model_selection}. In summary, the BoB Robust Zooming algorithm is more efficient and easier to use in practice, while yielding worse regret bound in theory. However, due to its practical applicability, we will implement this BoB Robust Zooming algorithm in the experiments. It is also noteworthy that we can attain a better regret bound with Algorithm~\ref{alg:rmel} under the weak adversary as shown in Theorem~\ref{thm:rmel}, which aligns with our expectation since the strong adversary considered here is more malicious and difficult to defend against.

%% file: exp.tex
\section{Experiments}\label{sec:exp}
\begin{figure}[t]
\begin{minipage}[b]{0.33\linewidth}
    \centering
    \includegraphics[width = 0.99\textwidth]{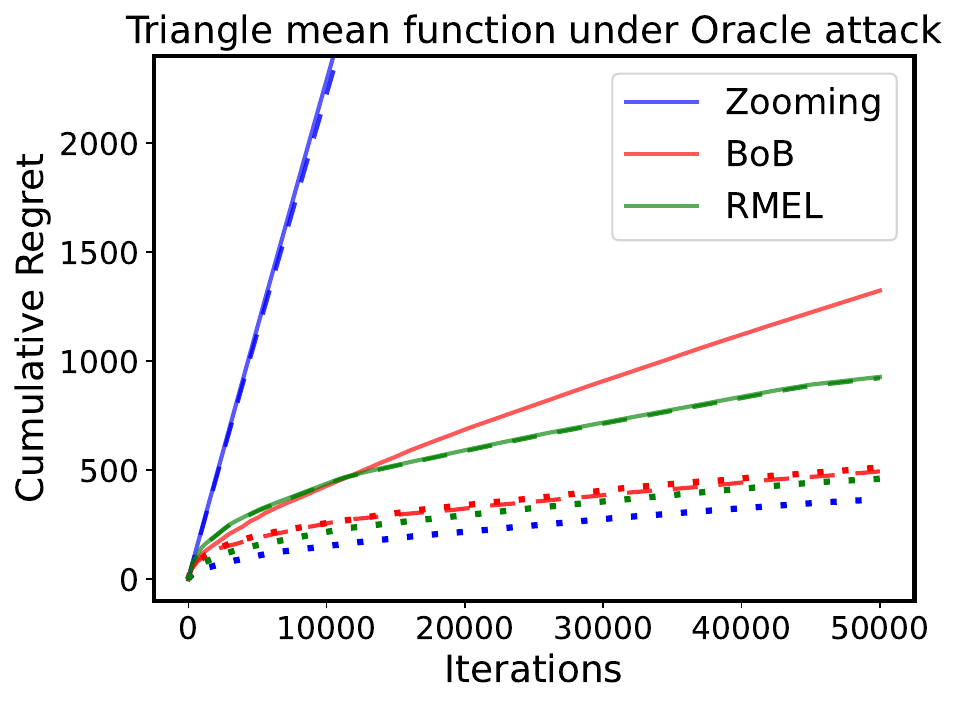}
\end{minipage}
\begin{minipage}[b]{0.33\linewidth}
    \centering
    \includegraphics[width = 0.98\textwidth]{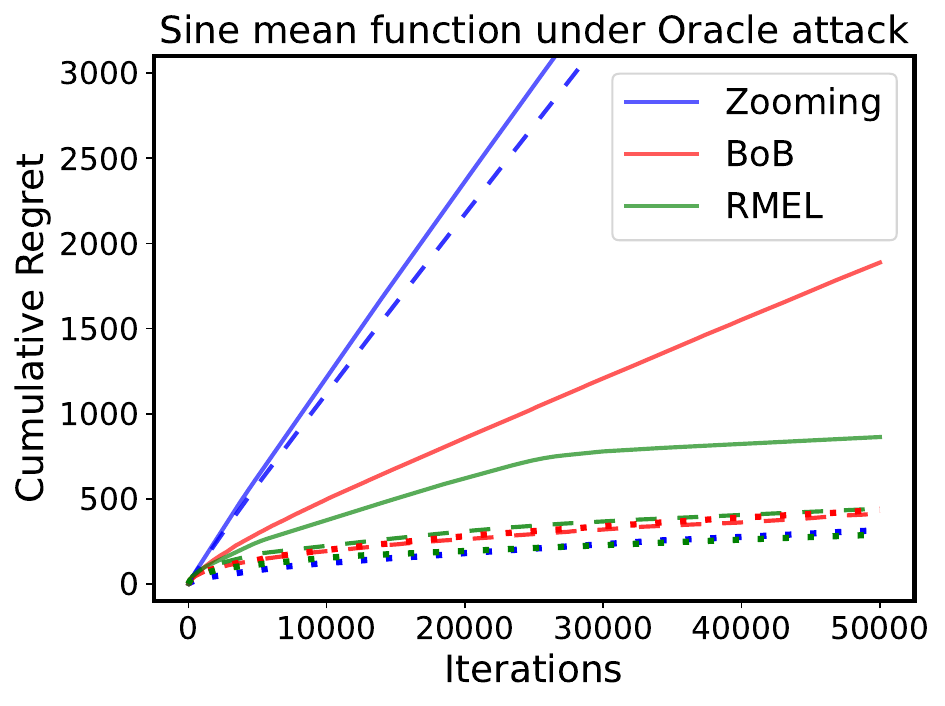}
\end{minipage}
\begin{minipage}[b]{0.33\linewidth}
    \centering
    \includegraphics[width = 0.995\textwidth]{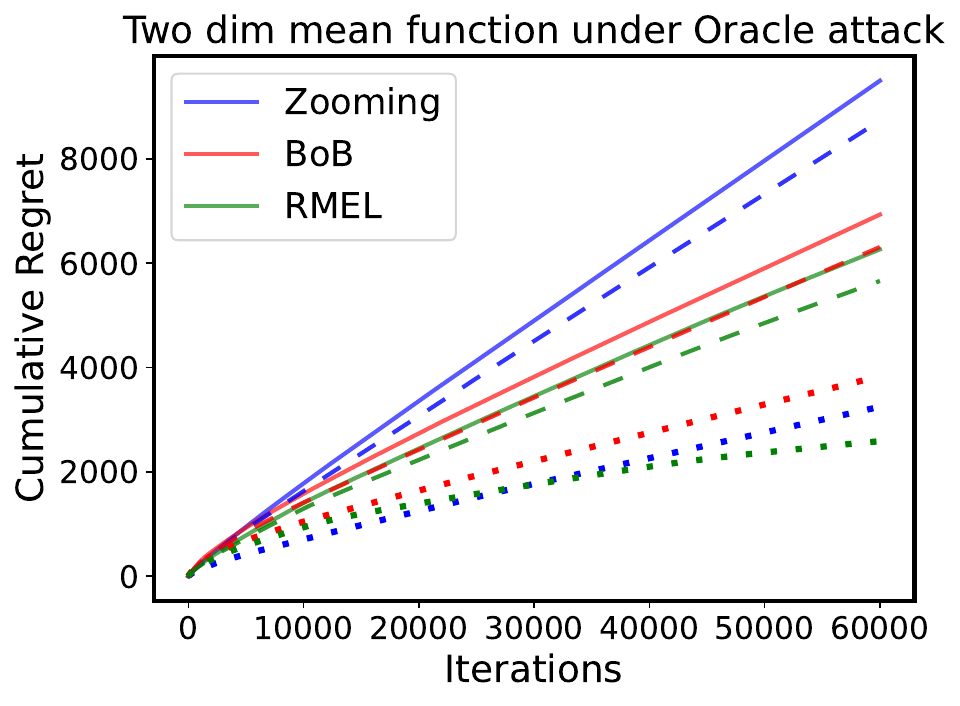}
\end{minipage}
\begin{minipage}[b]{0.33\linewidth}
    \centering
    \includegraphics[width = 0.99\textwidth]{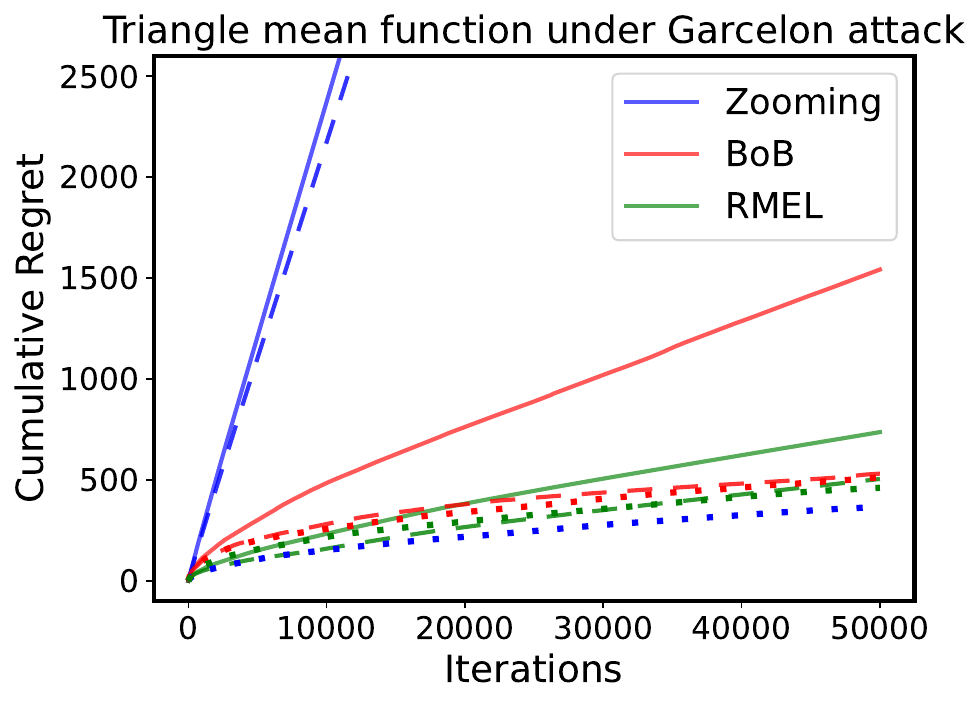}
\end{minipage}
\begin{minipage}[b]{0.33\linewidth}
    \centering
    \includegraphics[width = 0.96\textwidth]{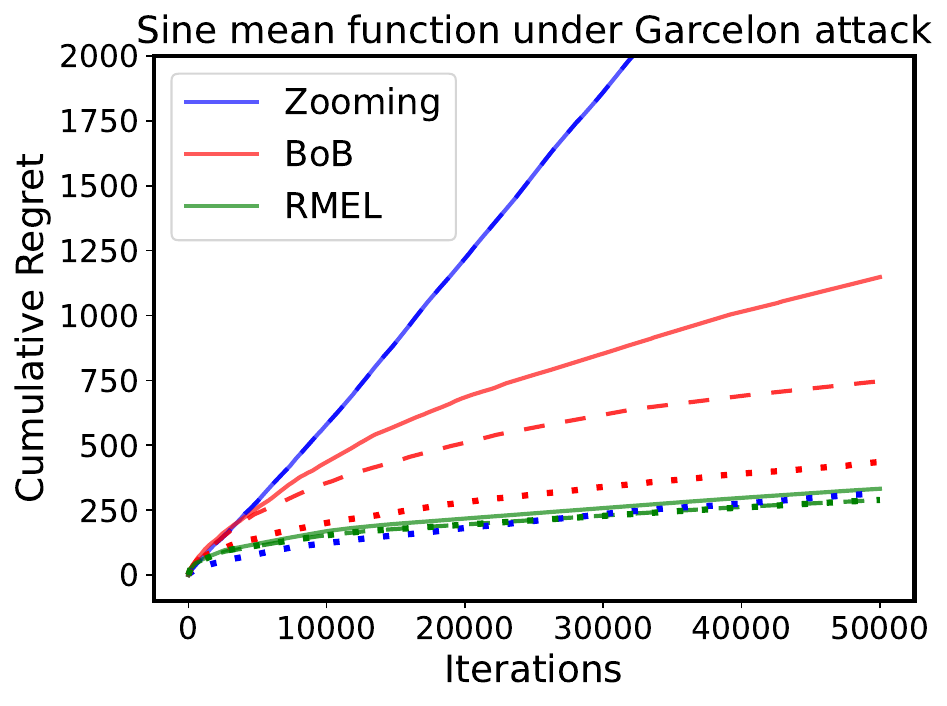}
\end{minipage}
\begin{minipage}[b]{0.33\linewidth}
    \centering
    \includegraphics[width = 0.995\textwidth]{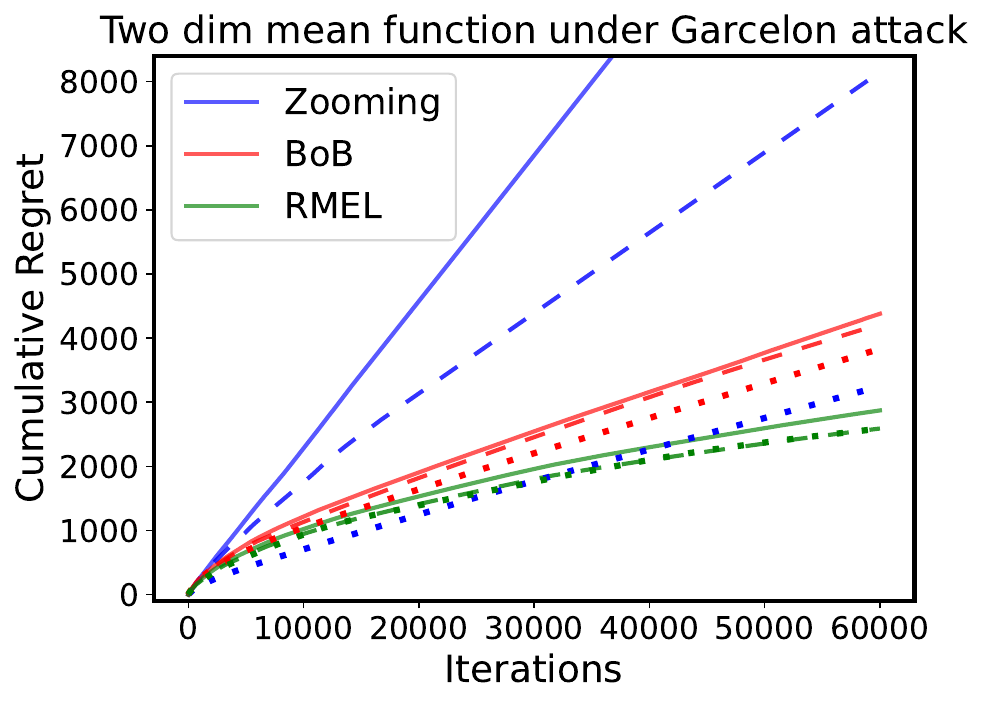}
\end{minipage}
\vspace{-3.2mm}
\caption{Plots of regrets of Zooming algorithm (blue), RMEL (green) and BoB Robust Zooming algorithm (red) under different settings with three levels of corruptions: (1) dotted line: no corruption; (2) dashed line: moderate corruptions; (3) solid line: strong corruptions. Numerical values of final cumulative regrets in our experiments are also displayed in Table~\ref{tab:exp} in Appendix~\ref{app:exp}.}
\label{plt:exp}
\vspace{-2.4mm}
\end{figure}
In this section, we show by simulations that our proposed RMEL and BoB Robust Zooming algorithm outperform the classic Zooming algorithm in the presence of adversarial corruptions. 
To firmly validate the robustness of our proposed methods, we use three types of models and two sorts of attacks with different corruption levels. We first consider the metric space $([0,1],| \! \cdot \! |)$ with two  expected reward functions that behave differently around their maximum: (1). $\mu(x) = 0.9-0.95|x-1/3|$ (triangle) and (2). $\mu(x) = 2/(3\pi) \cdot \sin{(3\pi x/2)}$ (sine). We then utilize a more complicated metric space $([0,1]^2,\| \! \cdot \! \|_{\infty})$ with the expected reward function (3). $\mu(x) = 1 - 0.8\|x - (0.75,0.75)\|_2 - 0.4\|x - (0,1)\|_2$ (two dim). We set the time horizon $T = 50,000 \,(60,000)$ for the metric space with $\cd = 1 \, (2)$ and the false probability rate $\delta = 0.01$. The random noise at each round is sampled IID from $N(0,0.01)$. Average cumulative regrets over 20 repetitions are reported in Figure \ref{plt:exp}. 

Since adversarial attacks designed for stochastic Lipschitz bandits have never been studied, we extend two types of classic attacks, named Oracle~\cite{jun2018adversarial}  for the MAB and Garcelon~\cite{garcelon2020adversarial} for the linear bandit, to our setting. The details of these two attacks are summarized as follows:

\setlist{leftmargin=2.5mm, topsep = 1mm}
\begin{itemize}
    \item \textbf{Oracle:} This attack \cite{jun2018adversarial} was proposed for the traditional MAB, and it pushes the rewards of ``good arms'' to the very bottom. Specifically,
        we call an arm is benign if the distance between it and the optimal arm is no larger than $0.2$. And we inject this attack by pushing the expected reward of any benign arm below that of the worst arm with an additional margin of $0.1$ with probability $0.5$.
    \item \textbf{Garcelon:} We modify this type of attack studied in~\cite{garcelon2020adversarial} for linear bandit framework, which replaces expected rewards of arms outside some targeted region with IID Gaussian noise. For $d=1$, since the optimal arm is set to be $1/3$ for both triangle and sine payoff functions, we set the targeted arm interval as $[0.5,1]$. For $d=2$, since the optimal arm is close to $(0.75,0.75)$, we set the targeted region as $[0,0.5]^2$. Here we contaminate the stochastic reward if the pulled arm is not inside the target region by modifying it into a random Gaussian noise $N(0,0.01)$ with probability $0.5$.
\end{itemize}


We consider the strong adversary in experiments as both types of attack are injected only if the pulled arms lie in some specific regions. Note although we originally propose RMEL algorithm for the weak adversary in theory, empirically we find it works exceptionally well (Figure~\ref{plt:exp}) across all settings here. We also conduct simulations based on the weak adversary and defer their settings and results to Appendix~\ref{app:exp} due to the limited space. The first Oracle attack is considered to be more malicious in the sense that it specifically focuses on the arms with good rewards, while the second Garcelon attack could corrupt rewards generated from broader regions, which may contain some ``bad arms'' as well.

Since there is no existing robust Lipschitz bandit algorithm, we use the classic Zooming algorithm~\cite{kleinberg2019bandits} as the baseline. As shown in Figure \ref{plt:exp}, we consider three levels of quantities of corruptions applied on each case to show how attacks progressively disturb different methods. Specifically, we set $C=0$ for the non-corrupted case, $C=3,000$ for the moderate-corrupted case and $C=4,500$ for the strong-corrupted case. Due to space limit, we defer detailed settings of algorithms to Appendix \ref{app:exp}.

From the plots in Figure~\ref{plt:exp}, we observe that our proposed algorithms consistently outperform the Zooming algorithm and achieve sub-linear cumulative regrets under both types of attacks, whereas the Zooming algorithm becomes incompetent and suffers from linear regrets even under a moderate volume of corruption. This fact also implies that the two types of adversarial corruptions used here are severely detrimental to the performance of stochastic Lipschitz bandit algorithms. And it is evident our proposed RMEL yields the most robust results under various scenarios with different volumes of attacks. It is also worth noting that the Zooming algorithm attains promising regrets under a purely stochastic setting, while it experiences a huge increase in regrets after the corruptions emerge. This phenomenon aligns with our expectation and highlights the fact that our proposed algorithms balance the tradeoff between accuracy and robustness in a much smoother fashion.


%% file: conclusion.tex
\section{Conclusion}\label{sec:conclu}
In this work we introduce a new problem of Lipschitz bandits in the presence of adversarial corruptions, and we originally provide efficient algorithms against both weak adversaries and strong adversaries when agnostic to the total corruption budget $C$. The robustness and efficiency of our proposed algorithms is then validated under comprehensive experiments. 

\textbf{Limitations:} For both the weak and strong adversary, our work leaves a regret gap between the lower bound deduced in Theorem~\ref{thm:lowerboundw} (Theorem~\ref{thm:lowerbounds}) and the upper bound in Theorem~\ref{thm:rmel} (Theorem~\ref{thm:model_selection}) when agnostic to $C$. Closing or narrowing this regret gap seems highly non-trivial since the regret gap still exists under the simpler MAB setting~\cite{gupta2019better}, and we will leave it as a future work.




%% file: app_thmrobust.tex
\subsection{Analysis of Theorem \ref{thm:robust}}\label{app:thmrobust}
We modify the proof in~\cite{kleinberg2019bandits} by dividing the cumulative regret into two parts, where the first part controls the error coming from the stochastic rewards and the second part deals with the extra error from adversarial corruptions in the following Appendix~\ref{app:sub:thmrobust}. In the beginning we will present some auxiliary lemmas for preparation.
\subsubsection{Useful Lemmas}
\begin{definition} 
We call it a clean process for Algorithm~\ref{alg:robust}, if for each time $t \in [T]$ and each active arm $v \in \X$ at any time $t$, we have $\lvert f(v) - \mu(v) \rvert \leq r(v)$. 
\end{definition}

Here we expand some notations from Algorithm~\ref{alg:robust}: we denote $n_t(v)$ as the number of times the arm $v$ has been pulled until the round $t$, and $f_t(x),r_t(x)$ as the corresponding average stochastic rewards and confidence radius respectively at time $t$ such that,
$$r_t(x) = \sqrt{\frac{4 \ln{(T)} + 2 \ln{(2/\delta)}}{n_t(x)}} + \frac{C}{n_t(x)}.$$
Note in our Algorithm~\ref{alg:robust} we do not write this subscript $t$ for these components since there is no ambiguity in the description. And W.l.o.g we assume the optimal arm $x_* = \arg \max_{x \in \X} \mu(x)$ is unique in $\X$.

\begin{lemma} \label{lem:clean}
Given the adversarial corruptions are at most $C$, for Algorithm~\ref{alg:robust}, the probability of a clean process is at least $1-\delta$. 
\end{lemma}
\proof
For each time $t \in [T]$, consider an arm $x \in \X$ that is active by the end of time $t$. Recall that when Algorithm~\ref{alg:robust} pulls the arm $x$, the reward is sampled IID from some unknown distribution $\mathbb{P}_x$ with expectation $\mu(x)$. And in the meanwhile, the stochastic reward may be corrupted by the adversary. Define random variables $U_{x,s}$ and values $C_{x,s}$ for $1 \leq s \leq n_t(x)$ as follows: for $s \leq n_t(x)$, $U_{x,s}$ is the stochastic reward from the $s$-th time arm $x$ is played and $C_{x,s}$ is the corruption injected on $U_{x,s}$ before the agent observing it. By applying Bernstein's Inequality, it naturally holds that
\begin{align*}
    &P\left( \left \lvert f_t(x) - \mu(x) \right \rvert \geq r_t(x) \right) = P\left( \left \lvert f_t(x) - \mu(x) \right \rvert \geq \sqrt{\frac{4 \ln{T}+2\ln{(2/\delta)}}{n_t(x)}} + \frac{C}{n_t(x)}\right ) \\
    = &P\left( \left \lvert \sum_{s=1}^{n_t(x)} \frac{U_{x,s}}{n_t(x)}+ \sum_{s=1}^{n_t(x)} \frac{C_{x,s}}{n_t(x)} - \mu(x) \right \rvert \geq \sqrt{\frac{4 \ln{T}+2\ln{(2/\delta)}}{n_t(x)}} + \frac{C}{n_t(x)}\right ) \\
    \leq & P\left( \left \lvert \sum_{s=1}^{n_t(x)} \frac{U_{x,s}}{n_t(x)} - \mu(x) \right \rvert + \sum_{s=1}^{n_t(x)} \frac{\left\lvert C_{x,s}\right\rvert}{n_t(x)} \geq \sqrt{\frac{4 \ln{T}+2\ln{(2/\delta)}}{n_t(x)}} + \frac{C}{n_t(x)}\right ) \\
    \stackrel{\textup{(i)}}{\leq} & P\left( \left \lvert \sum_{s=1}^{n_t(x)} \frac{U_{x,s}}{n_t(x)} - \mu(x) \right \rvert  \geq \sqrt{\frac{4 \ln{T}+2\ln{(2/\delta)}}{n_t(x)}} \right ) 
    \leq  2\cdot \exp\left(-\frac{n_t(x)}{2} \times \frac{4\ln{T}+2\ln{(2/\delta)}}{n_t(x)}\right) \\
    & \qquad \qquad \qquad \qquad \qquad \qquad \qquad \qquad \qquad \qquad \; \;= \delta T^{-2},
\end{align*}
where the inequality (i) comes from the fact that the total corruption budget is at most $C$. Since there are at most $t$ active arms by time $t$, by taking the union bound over all active arms it holds that,
$$P\left(\forall \, \text{active arm $x$ at round $t$}, \left \lvert f_t(x) - \mu(x) \right \rvert \leq r_t(x) \right) \geq 1 - \delta T^{-1}, \quad \forall t \in [T].$$
Finally, we take the union bound over all round $t \leq T$, and it holds that,
$$P\left(\forall t \leq T, \forall \, \text{active arm $x$ at round $t$}, \left \lvert f_t(x) - \mu(x) \right \rvert \leq r_t(x) \right) \geq 1 - \delta T^{-1},$$
which implies that the probability of a clean process is at least $1-\delta$.
\hfill \qedsymbol

\begin{lemma}\label{lem:notremove}
If it is a clean process and the optimal arm $x_* \in B(v,r_t(v))$, then $\B(v,r_t(v))$ could never be eliminated from Algorithm \ref{alg:robust} for any $t\in[T]$ and active arm $v$ at round $t$.
\end{lemma}
\proof
Recall that from Algorithm \ref{alg:robust}, at round $t$ the ball $\B(u,r_t(u))$ would be discarded if we have for some active arm $v$ s.t.
$$f_t(v) -r_t(v) > f_t(u) + 2 r_t(u).$$
If $x_* \in \B(u,r_t(u))$, then it holds that 
$$f_t(u) + 2 r_t(u) \stackrel{\textup{(i)}}{\geq} \mu(u) + r_t(u) \geq \mu(u)+D({u},{x_*}) \stackrel{\textup{(ii)}}{\geq} \mu(x_*),$$
where inequality (i) is due to the clean process and inequality (ii) comes from the fact that $\mu(\cdot)$ is a Lipschitz function. On the other hand, we have that for any active arm $v$,
$$\mu(v) \geq f_t(v) - r_t(v), \quad \mu(x_*) \geq \mu(v).$$
Therefore, it naturally holds that
$$f_t(v) -r_t(v) \leq f_t(u) + 2 r_t(u).$$
\hfill \qedsymbol

\begin{lemma}\label{lem:distance>delta}
If it is a clean process, then for any time $t$ and any (previously) active arm $v$ we have $\Delta(v) \leq 3 r_t(v)$. Furthermore, we could deduce that $D(u,v) \geq \min\{\Delta(u), \Delta(v)\}/3$ for any pair of (previously) active arms $(u,v)$ by the time horizon $T$.
\end{lemma}
\proof
Let $S_t$ be the set of all arms that are active or were once active at round $t$. Suppose an arm $x_t$ is played at time $t$. If $x_t$ is just played for one time, i.e. $x_t$ is just activated at time $t$, then we naturally have that,
$$\Delta(x_t) \leq 1 \leq 3 r_t(x_t),$$
since the diameter of $\X$ is at most 1. Otherwise, if $x_t$ was played before, i.e. $x_t$ is chosen based on the selection rule instead of the activation rule, we will claim that 
$$\mu(x_*) \leq f_t(x_t) + 2r_t(x_t) \leq \mu(x_t) + 3r_t(x_t),$$
under a clean process. First we will show that $f_t(x_t) + 2r_t(x_t) \geq \mu(x_*)$. Recall that the optimal arm $x_*$ is never eliminated according to \ref{lem:notremove} under a clean process and hence is covered by some confidence ball, i.e. $x_* \in \B(x', r_t(x')), \exists x' \in S_t$. Then based on the selection rule, it holds that
$$f_t(x_t) + 2r_t(x_t) \geq f_t(x') + 2r_t(x') \geq \mu(x')+r_t(x') \geq \mu(x_*)+r_t(x') - D(x_*,x') \geq \mu(x_*).$$
On the other hand, it holds that,
$$f_t(x_t) + 2r_t(x_t) \leq \mu(x_t) + 3r_t(x_t)$$
since it is a clean process. And these two results directly imply that 
\begin{align}
    \mu(x_*)- \mu(x_t) = \Delta(x_t) \leq 3r_t(x_t). \label{eq:delta<r}
\end{align}
For the other active arms $v \in S_t$ that was played before time $t$, let $s < t$ be the last time arm $v$ was played, where we have $f_t(v) = f_s(v)$ and $r_t(v) = r_s(v)$, and then based on Eqn.~\eqref{eq:delta<r} it holds that $\Delta(v) \leq 3 r_s(v) = 3 r_t(v)$. 

Furthermore, we will show that $D(u,v) \geq \min\{\Delta(u), \Delta(v)\}/3$ for any pair of active arms $(u,v)$ by the time horizon $T$. W.l.o.g we assume that $v$ was activated before $u$, and $u$ was first activated at some time $s'$. Then if $v$ was active at the time $s'$ it naturally holds that $D(u,v) > r_{s'}(v) \geq \Delta(v)/3$ according to the activation rule. If $v$ was removed at the time $s'$ then we also have $D(u,v) > r_{s'}(v)$ since $u$ was not among the discarded region, and hence $D(u,v) \geq \Delta(v)/3$ holds as well. And this concludes our proof.
\hfill \qedsymbol
\subsubsection{Proof of Theorem \ref{thm:robust}} \label{app:sub:thmrobust}
We modify the original argument for Zooming algorithm~\cite{kleinberg2019bandits} to decently resolve the presence of adversarial corruptions. In summary, we could bound the cumulative regret of order $\tilde O \left(T^{\frac{\zd+1}{\zd+2}} + C^{\frac{1}{\zd+1}} T^{\frac{\zd}{\zd+1}}\right)$: the first term is the regret caused by the stochastic rewards, which is identical to the regret we have without any corruptions; the second quantity bounds the additional regret caused by the corruptions.

Denote $S_T$ as the active (or previously active) arm set across the time horizon $T$. Then based on Lemma \ref{lem:distance>delta}, for any $x \in S_T$ it holds that,
\begin{align}
    \Delta(x) \leq 3 r_T(x) = 3 \sqrt{\frac{4 \ln{(T)} + 2 \ln{(2/\delta)}}{n_T(x)}} + \frac{3C}{n_T(x)}. \nonumber
\end{align}
And this indicates that
\begin{align}
    \Delta(x)n_T(x) \leq 3 \sqrt{\left({4 \ln{(T)} + 2 \ln{\left(\frac{2}{\delta} \right)}}\right){n_T(x)}} + {3C}. \label{eq:thm1_1}
\end{align}
Then we denote
$$B_{i,T} = \left\{v \in S_T \, : \, 2^{i} \leq \frac{1}{\Delta(v)} < 2^{i+1} \right\}, \quad \text{where } \, S_T = \bigcup_{i=0}^{+\infty} B_{i,T},$$
and write $r_i = 2^{-i}$. Then for arbitrary $u,v \in B_{i,T}, i \geq 0$, we have
$$\frac{r_i}{2} < \Delta(u) \leq r_i, \quad \frac{r_i}{2} < \Delta(v) \leq r_i,$$
which implies that $D(x,y) > r_i/6$ under a clean process based on Lemma \ref{lem:distance>delta}. Based on the definition of the zooming dimension $\dz$, it follows that $\vert B_{i,T} \vert \leq O(r_i^{\dz})$. Subsequently, for any $0<\rho < 1$ it holds that
\begin{align}
\sum_{\substack{v \in S_T, \\ \Delta(v) > \rho}} 1 \leq \sum_{i < -\log_2(\rho)}O(r_i^{-\zd}) = O\left( \frac{1}{\rho^{\zd}} \right). \label{eq:strho}
\end{align}
Now we define the set $I$ as:
$$I \coloneqq \left\{v \in S_T : C \leq \sqrt{\left({4 \ln{(T)} + 2 \ln{\left(\frac{2}{\delta} \right)}}\right){n_T(v)}}  \right\}.$$
When an arm $v$ is in the set $I$, the cumulative regret in terms of it would be more related to the stochastic errors other than the adversarial attacks. Subsequently, we could divide the cumulative regret into two quantities:
\begin{align}
    Re&gret_T = \sum_{v \in S_T} \Delta(v)n_T(v) = \sum_{v \in S_T \cap I} \Delta(v)n_T(v)+ \sum_{v \in S_T \cap I^c} \Delta(v)n_T(v) \nonumber \\
    &= \sum_{\substack{v \in S_T \cap I, \\ \Delta(v) \leq \rho_1}} \Delta(v)n_T(v)+ \sum_{\substack{v \in S_T \cap I, \\ \Delta(v) > \rho_1}} \Delta(v)n_T(v) +  \sum_{\substack{v \in S_T \cap I^c, \\ \Delta(v) \leq \rho_2}} \Delta(v)n_T(v)+ \sum_{\substack{v \in S_T \cap I^c, \\ \Delta(v) > \rho_2}} \Delta(v)n_T(v) \nonumber \\
    &\stackrel{\textup{(i)}}{\leq} \rho_1 T + 2 \sum_{\substack{v \in S_T \cap I, \\ \Delta(v) > \rho_1}} 3 \sqrt{\left({4 \ln{(T)} + 2 \ln{\left(\frac{2}{\delta} \right)}}\right){n_T(v)}} + \rho_2 T + 2\sum_{\substack{v \in S_T \cap I^c, \\ \Delta(v) > \rho_2}} 3C \nonumber \\
    &\stackrel{\textup{(ii)}}{\lesssim} \rho_1 T + \sqrt{\ln{\left(\frac{T}{\delta}\right)}} \sqrt{\left( \sum_{\substack{v \in S_T \cap I, \\ \Delta(v) > \rho_1}} n_T(v)\right) \left(\sum_{\substack{v \in S_T \cap I, \\ \Delta(v) > \rho_1}} 1 \right)} +  \rho_2 T + C  \sum_{\substack{v \in S_T \cap I^c, \\ \Delta(v) > \rho_2}} 1 \nonumber \\
    &\lesssim \rho_1 T + \sqrt{\ln{\left(\frac{T}{\delta}\right)}} \sqrt{\left( \sum_{\substack{v \in S_T \cap I, \\ \Delta(v) > \rho_1}} n_T(v)\right) \left(\sum_{\substack{v \in S_T \cap I, \\ \Delta(v) > \rho_1}} 1 \right)} +  \rho_2 T + C  \sum_{\substack{v \in S_T \cap I^c, \\ \Delta(v) > \rho_2}} 1 \nonumber \\
    &\stackrel{\textup{(iii)}}{\lesssim} \rho_1 T + \sqrt{\ln{\left(\frac{T}{\delta}\right)}} \sqrt{T} \left(\frac{1}{\rho_1}\right)^{\frac{\zd}{2}} + \rho_2 T + C \left(\frac{1}{\rho_2}\right)^{\zd} \label{eq:thm1final}.
\end{align}
The inequality (i) comes from the definition of set $I$ and Eqn. \eqref{eq:thm1_1}, and inequality (ii) is due to the Cauchy-Schwarz inequality where $\lesssim$ denotes ``less in order''. Furthermore, we get inequality (iii) based on Eqn. \eqref{eq:strho}. Note Eqn. \eqref{eq:thm1final} holds for arbitrary $\rho_1, \rho_2 \in (0,1)$, and hence by taking
$$\rho_1 = T^{-\frac{1}{\zd+2}} \ln{(T)}^{\frac{1}{\zd+2}}, \quad \rho_2 = T^{-\frac{1}{\zd+1}}C^{\frac{1}{\zd+1}},$$ 
we have
$$Regret_T  = O \left(\ln{(T)}^{\frac{1}{\zd+2}} T^{\frac{\zd+1}{\zd+2}} + C^{\frac{1}{\zd+1}} T^{\frac{\zd}{\zd+1}} \right) = \tilde O \left(T^{\frac{\zd+1}{\zd+2}} + C^{\frac{1}{\zd+1}} T^{\frac{\zd}{\zd+1}} \right). $$
And this concludes our proof.
\hfill \qedsymbol

\begin{remark}\label{rem:thmrobust}
Note we could replace the second term of $r(x)$ with $\min\{1,C/n(x)\}$, i.e.
$$r_t(x) = \sqrt{\frac{4 \ln{(T)} + 2 \ln{(2/\delta)}}{n_t(x)}} + \min\left\{1,\frac{C}{n_t(x)}\right\},$$
since we know each instance of attack is assumed to be upper bounded by $1$. And all our analyses and Lemmas introduced above could be easily verified. Specifically, the core Lemma~\ref{lem:clean} still holds as
$$\sum_{s=1}^{n_t(x)} \frac{C_{x,s}}{n_t(x)} \leq \sum_{s=1}^{n_t(x)} \frac{1}{n_t(x)}  = 1.$$
\end{remark}

%% file: app_thmrmel.tex
\subsection{Analysis of Theorem \ref{thm:rmel}}\label{app:thmrmel}
\subsubsection{Useful Lemmas}
We first present supportive Lemmas and some of them are adapted from the results in~\cite{feng2022lipschitz,lykouris2018stochastic}.
\begin{lemma}\label{lem:bernoulli} 
For a sequence of IID Bernoulli trials with a fix success probability $p$, then with probability $1-\delta$, we could at most observe $\left[(1-p)\ln{(1/\delta)}/p\right]$ failures until the first success.
\end{lemma}
\proof
This is based on the property of negative binomial distribution: after we complete the first $N$ trials, the probability of no success is $(1-p)^N$. To ensure this value is less than $\delta$, we get
$$N = \log_{1-p}(\delta) = \frac{\ln{(1/\delta)}}{\ln{\left(1/(1-p)\right)}} = \frac{\ln{(1/\delta)}}{\ln{\left(1+p/(1-p)\right)}}.$$
By using the inequality $\ln{(x+1)} \leq x, \forall x>-1$, we could take $N = \left[(1-p)\ln{(1/\delta)}/p\right]$.
\hfill\qedsymbol
\begin{lemma}\label{lem:boundc}
\textup{(Adapted from Lemma 3.3~\cite{lykouris2018stochastic})} In Algorithm~\ref{alg:rmel}, for any layer whose tolerance level exceeds the unknown $C$, i.e. any layer with index $i \in [l^*]$ s.t. $v_i \geq C$, with probability at least $1-\delta$, this layer suffers from at most corruptions of amount $\left(\ln{(1/\delta)}+2e-1\right)$.
\end{lemma}
\proof 
The proof of this Lemma is an adaptation from the proof of Lemma 3.3 in~\cite{lykouris2018stochastic}, and we present the detailed proof here for completeness: 

In the beginning, we introduce an important result (Lemma 1 in~\cite{beygelzimer2011contextual}): Let $X_1,\dots,X_T$ be a real-valued martingale difference sequence, i.e. $\forall t\in [T], \mathbb{E}(X_t|X_{t-1},\dots,X_1) = 0$. And $X_t \leq R$. Denote $V = \sum_{t=1}^T \mathbb{E}(X_t^2|X_{t-1},\dots,X_1)$. Then for any $\delta > 0$, it holds that,
$$P\left( \sum_{t=1}^T X_t > R \ln{\left(\frac{1}{\delta} \right)} + \frac{e-2}{R} \cdot V \right) \leq \delta.$$
Assume a layer whose tolerance level $\tilde C$ is no less than $C$, and hence the probability of pulling this layer would be $1/\tilde C \leq 1/C$. For this layer, let $\tilde{C}_x^t$ be the corruption that is observed at round $t$ when arm $x$ is pulled, $x \in X$. Then at any time $t$, if the adversary selects corruption $c_t(a)$ then we know $\tilde{C}_x^t$ is equal to $c_t(a)$ with probability $1/\tilde C$ and $0$ otherwise. Denote the filtration $\tilde{\mathcal{F}}_t$ containing all the realizations of random variables before time $t$. And hence at time $t$ the adversary could contaminate the stochastic rewards of $\X$ according to $\tilde{\mathcal{F}}_t$. Let $\tilde a_t$ be the arm that would be selected if this layer is chosen at the time $t$. Since our Algorithm~\ref{alg:rmel} is deterministic in terms of the active region conditioned on selecting each layer, and the pulled arm is randomly selected from the active region. Therefore, the selection of $\tilde a_t$ is also independent with $\tilde{C}_x^t$ given $\tilde{\mathcal{F}}_t$. We construct the martingale as:
$$X_t = \left|\tilde{C}_x^t\right| - \mathbb{E} \left(\left| \tilde{C}_x^t \right|  \, \big| \, \tilde{\mathcal{F}}_t \right).$$
Therefore, it holds that
\begin{align*}
    \mathbb{E}(X_t^2 | X_{t-1},\dots,X_1) = \frac{1}{\tilde C} \left(\left|c_t(a)\right| - \frac{\left|c_t(a)\right|}{\tilde C} \right)^2 + \frac{\tilde{C}-1}{\tilde C} \left(\frac{\left|c_t(a)\right|}{\tilde C} \right)^2 \leq 2 \frac{\left|c_t(a)\right|}{C},
\end{align*}
since we have that $C \leq \tilde{C}$ and $\left|c_t(a)\right| \leq 1$. And conclusively it holds that 
$$V = \sum_{t=1}^T \mathbb{E}(X_t^2|X_{t-1},\dots,X_1) \leq \sum_{t=1}^T 2 \frac{\left|c_t(a)\right|}{C} \leq 2.$$
Furthermore, it naturally holds that $X_t \leq 1$ due to the fact that $\left|c_t(a)\right| \leq 1$. Based on Lemma 1 in~\cite{beygelzimer2011contextual} we introduced above, with probability at least $1-\delta$, it holds that 
$$\sum_{t=1}^T X_t \leq \ln{\left(\frac{1}{\delta}\right)} + 2(e-2).$$
On the other hand, we can trivially deduce that the expected corruption injected in this layer is at most $1$ since we have total amount of corruptions $C$ and the probability of choosing this layer at each time is fixed as $1/\tilde C \leq 1/C$. Conclusively, we have with probability at least $1-\delta$,
$$\sum_{t=1}^T \left|\tilde{C}_x^t\right| = \sum_{t=1}^T X_t + \mathbb{E} \left( \sum_{t=1}^T \left| \tilde{C}_x^t \right|  \, \big| \, \tilde{\mathcal{F}}_t \right) \leq \ln{\left(\frac{1}{\delta}\right)} + 2(e-2) +1 = \ln{\left(\frac{1}{\delta}\right)} + 2e -1.$$
And this completes the proof.
\hfill \qedsymbol
\begin{definition}\label{def:cleanrmel} 
We call it a clean process for Algorithm~\ref{alg:rmel}, if for any time $t \in [T]$, any layer $l \in [l^*]$ whose tolerance level $v_l \geq C$, any active region $A \in \A_l$ and any $x \in A$ at time $t$, we have 
$$\lvert f_{l,A} - \mu(x) \rvert \leq \frac{1}{2^{m_l}} + \sqrt{\frac{4 \ln{(T)} + 2 \ln{(4/\delta)}}{n_{l,A}}} + \frac{\ln{(T)} + \ln{(4/\delta)}}{n_{l,A}}$$
hold for some $0<\delta<1$.
\end{definition}
To facilitate our analysis in the rest of this section, we expand notations here for Algorithm~\ref{alg:rmel}. Similar as in Appendix~\ref{app:thmrobust}, we would add the subscript time $t$ to some notations used in Algorithm~\ref{alg:rmel}.
\begin{itemize*}
    \item $m_{l,t}$: epoch index of layer $l$ at time $t$;
    \item $n_{l,t}$: number of selecting the layer $l$ at time $t$ since the last refresh (line 10 of Algorithm~\ref{alg:rmel}) on the layer $l$;
    \item $\A_{l,t}$: active arm set of layer $l$ at time $t$;
    \item $n_{l,A,t}$: number of selecting the layer $l$ and active region $A\in \A_{l,t}$ by time $t$ since the last refresh on the layer $l$;
    \item $f_{l,A,t}$: average stochastic rewards of selecting the layer $l$ and active region $A\in \A_{l,t}$ by time $t$ since the last refresh on the layer $l$.
\end{itemize*}
We also denote $l_0$ as the minimum index of layer whose tolerance level just surpasses $C$, i.e. $l_0 = \arg \min \{l \in [l^*]: v_l \geq C\}$. Therefore, we get a clean process defined in Definition~\ref{def:cleanrmel} \textit{iff.} the following set $\Phi$ holds:
\begin{align}
    \Phi = \Bigg\{\left\lvert f_{l,A,t} - \mu(x) \right\rvert \leq\frac{1}{2^{m_l}} + & \sqrt{\frac{4 \ln{(T)} + 2 \ln{(4/\delta)}}{n_{l,A,t}}} + \frac{\ln{(T)} + \ln{(4/\delta)}}{n_{l,A,t}} : \nonumber
    \\& \forall x \in A, \, \forall A \in \A_{l,t}, \, \forall l \in \{l_0, l_0+1, \dots, l^*\}, \, \forall t \in [T] \Bigg\}. \label{eq:phiset}
\end{align}
Note the following Lemmas hold for either Algorithm~\ref{alg:rmel} or its variant Algorithm~\ref{alg:alterrmel} since we define the clean process with respect to each round $t$ instead of each epoch. Note we only need to prove the set $\Phi$ holds at the end of each epoch for the analysis of Algorithm~\ref{alg:rmel}. W.l.o.g. we will just prove the regret bound in Theorem~\ref{thm:rmel} of Algorithm~\ref{alg:rmel}, while it is easy to verify that the same arguments and Theorem~\ref{thm:rmel} hold for Algorithm~\ref{alg:alterrmel} as well.
\begin{corollary}\label{coro:pull}
With probability at least $1-\frac{\delta}{4}$, we select one time of layer $l_0$ at most every $BC \log(4T/\delta)$ times of other layers simultaneously. 
\proof
The proof is straight forward based on Lemma~\ref{lem:bernoulli}. According to the construction of $\{v_l\}_{l=1}^{l^*}$, it holds that $C \leq v_{l_0} < BC$. This implies that the probability of sampling layer $l_0$ at each round is at least $\frac{1}{BC}$. Therefore, after sampling layer $l_0$ in line 2 of Algorithm~\ref{alg:rmel}, with probability at least $1-\frac{\delta}{4T}$, we would sample all the other layers for at most 
$$BC \frac{\log(4T/\delta)}{1-\frac{1}{BC}} \leq BC \,{\log(4T/\delta)}$$
times. Since we know the number of time sampling layer $l_0$ is naturally at most $T$, by taking the union bound, we conclude the proof of Corollary \ref{coro:pull}.
\hfill \qedsymbol
\end{corollary}
\begin{lemma}\label{lem:cleanrmel}
For algorithm~\ref{alg:rmel}, the probability of a clean process is at least $1-\frac{3}{4}\delta$, i.e. $P(\Phi) \geq 1 - \frac{3}{4}\delta$.
\end{lemma}
\proof 
For each layer $l$ whose tolerance level surpasses $C$, i.e. $l \geq l_0$, we know the probability of sampling this layer in line 2 of Algorithm~\ref{alg:rmel} is at most ${1}/{C}$, and this indicates that with probability at least $1-\delta_1$, this layer suffers from at most $\left( -\ln{(\delta_1)} +2e-1 \right)$ levels of corruptions based on Lemma~\ref{lem:boundc}. Note the number of layers is less than $\log_B(T)$. This indicates that by taking the union bound on all layers whose tolerance levels surpass $C$, we have with probability at least $1-\delta_1$, all these layers suffer from at most $\left( \ln{\left(\frac{\log_B(T)}{\delta_1}\right)} +2e-1\right)$ levels of corruptions across the time horizon $T$. And note
$$\ln{\left(\frac{\log_B(T)}{\delta_1}\right)} +2e-1 \leq \ln{\left(\frac{T}{\delta_1}\right)}$$
since it is natural to have $T/\log_B(T) \geq e^3$. Then for any time $t$, any layer $l \geq l_0$ and any active region $A \in \A_{l,t}$, define $x_{A,s}$, $C_{A,s}$ and random variables $U_{A,s}$ as the $s$-th time arm pulled, the stochastic reward from pulling $x_{A,s}$ and the corruption injected on $U_{A,s}$ for $1 \leq s \leq n_{l,A,t}$. Also
denote
$$r_{l,A,t} = \frac{1}{2^{m_{l,t}}} + \sqrt{\frac{4 \ln{(T)} + 2 \ln{(4/\delta)}}{n_{l,A,t}}} + \frac{\ln{(T)} + \ln{(4/\delta)}}{n_{l,A,t}}.$$
With probability at least $1-\delta/4$, from the above argument, we know that all layers with the index at least $l_0$ suffer from at most $\ln{\left(\frac{4T}{\delta}\right)}$ levels of corruptions across the time horizon $T$. Denote this event as $\Psi$, i.e. $P(\Psi) \geq 1 - \delta/4$, then under $\Psi$ it holds that
\begin{align*}
    &P\left( \lvert f_{l,A,t} - \mu(x) \rvert \leq r_{l,A,t}, \; \forall x \in A \right) \nonumber \\
    &=P\left( \left\lvert  \sum_{s=1}^{n_{l,A,t}} \frac{U_{x,s}}{n_{l,A,t}}+ \sum_{s=1}^{n_{l,A,t}} \frac{C_{x,s}}{n_{l,A,t}} - \mu(x) \right \rvert \leq r_{l,A,t}, \; \forall x \in A \right) \nonumber \\
    &\geq P\left( \left\lvert  \sum_{s=1}^{n_{l,A,t}} \frac{U_{x,s}}{n_{l,A,t}} \!- \!\sum_{s=1}^{n_{l,A,t}} \frac{\mu(x_{A,s})}{n_{l,A,t}} \right \rvert  \!+ \!\left\lvert  \sum_{s=1}^{n_{l,A,t}} \frac{\mu(x_{A,s})}{n_{l,A,t}} \!- \!\mu(x) \right \rvert \!+\! \left\lvert   \sum_{s=1}^{n_{l,A,t}} \frac{C_{x,s}}{n_{l,A,t}} \right \rvert \leq r_{l,A,t}, \forall x \in A \right) \nonumber \\
    &\stackrel{\textup{(i)}}{\geq}  \!P\!\left( \left\lvert  \sum_{s=1}^{n_{l,A,t}}\! \frac{U_{x,s}}{n_{l,A,t}}\! -\! \sum_{s=1}^{n_{l,A,t}}\! \frac{\mu(x_{A,s})}{n_{l,A,t}} \right \rvert \!  +\! \left\lvert  \sum_{s=1}^{n_{l,A,t}} \!\frac{\mu(x_{A,s})}{n_{l,A,t}} - \mu(x) \right \rvert  \leq \frac{1}{2^{m_{l,t}}} \!+ \!\sqrt{\frac{2 \ln{(4T^2/\delta)}}{n_{l,A,t}}}, \forall x \in A \!\right) \nonumber \\
    &\stackrel{\textup{(ii)}}{\geq} P\left( \left\lvert  \sum_{s=1}^{n_{l,A,t}} \frac{U_{x,s}}{n_{l,A,t}} - \sum_{s=1}^{n_{l,A,t}} \frac{\mu(x_{A,s})}{n_{l,A,t}} \right \rvert    \leq   \sqrt{\frac{2 \ln{(4T^2/\delta)}}{n_{l,A,t}}}\right)  \\ 
    &\geq 1-\frac{\delta}{2} \cdot T^{-2}.
\end{align*}
Inequality (i) is due to the definition of event $\Psi$ and inequality (ii) comes from the fact that the diameter of $A$ is at most $1/2^{m_{l,t}}$ and $\mu(\cdot)$ is a Lipschitz function.
We know that at most $T$ active regions would be played across time $T$. By taking the union bound on all rounds $t \in [T]$ and all active regions that have been played, it holds that 
$$P\left( \lvert f_{l,A,t} - \mu(x) \rvert \leq r_{l,A,t}, \, \forall x \in A, \, \forall A \in \A_{l,t}, \, \forall l \in \{l_0, l_0+1, \dots, l^*\}, \, \forall t \in [T]  \right) \geq 1 - \frac{\delta}{2}$$
under the event $\Psi$. Since $P(\Psi) \geq 1 - \delta/4$, overall it holds that
$$P\left( \lvert f_{l,A,t} - \mu(x) \rvert \leq r_{l,A,t}, \, \forall x \in A, \, \forall A \in \A_{l,t}, \, \forall l \in \{l_0, l_0+1, \dots, l^*\}, \, \forall t \in [T]  \right) \geq 1 - \frac{3\delta}{4},$$
i.e. $P(\Phi) \geq 1-{3\delta}/{4}$. And this concludes our proof.
\hfill \qedsymbol
\begin{lemma}\label{lem:rmel4r}
We have $r_{l,A,t} \leq {2}/{2^{m_{l,t}}}$ if $n_{l,A,t} = 6\ln(4T/\delta)\cdot 4^{m_{l,t}}$.
\end{lemma}
\proof
Based on the formulation of $r_{l,A,t}$
$$r_{l,A,t} = \frac{1}{2^{m_{l,t}}} + \sqrt{\frac{4 \ln{(T)} + 2 \ln{(4/\delta)}}{n_{l,A,t}}} + \frac{\ln{(T)} + \ln{(4/\delta)}}{n_{l,A,t}}.$$
It suffices to show that 
\begin{align}
\sqrt{\frac{4 \ln{(T)} + 2 \ln{(4/\delta)}}{n_{l,A,t}}} + \frac{\ln{(T)} + \ln{(4/\delta)}}{n_{l,A,t}} \leq \frac{1}{2^{m_{l,t}}} \label{eq:remllem1}
\end{align}
by taking $n_{l,A,t} = 6\ln(4T/\delta)\cdot 4^{m_{l,t}}$.
Firstly, we have that 
\begin{align*}
    \sqrt{\frac{4 \ln{(T)} + 2 \ln{(4/\delta)}}{n_{l,A,t}}} \leq 2\sqrt{\frac{\ln{(T)} + \ln{(4/\delta)}}{6\ln(4T/\delta)\cdot 4^{m_{l,t}}}}
    &\leq 2\sqrt{\frac{\ln{(T)} + \ln{(4/\delta)}}{(3+2\sqrt{2})\ln(4T/\delta)\cdot 4^{m_{l,t}}}} \\& \leq (2\sqrt{2}-2)  \frac{1}{2^{m_{l,t}}}
\end{align*}
Secondly, it holds that
\begin{align*}
     \frac{\ln{(T)} + \ln{(4/\delta)}}{n_{l,A,t}} \leq \frac{1}{3+2\sqrt{2}} \frac{1}{4^{m_{l,t}}} \leq (3-2\sqrt{2}) \frac{1}{2^{m_{l,t}}}.
\end{align*}
Combining the above two results, we have Eqn. \eqref{eq:remllem1} holds, which concludes our proof.
\hfill \qedsymbol
\begin{lemma}\label{lem:deltarmel}
Under a clean process, for any layer $l$ whose tolerance level $v_l$ is no less than $C$, i.e. $l \geq l_0$, it holds that
$$\Delta(x) \leq 16/2^{m_{l,t}}, \quad \forall x \in A, \forall A \in \A_{l,t}, \forall t \in [T].$$
\end{lemma}
\proof 
Here we will focus on Algorithm~\ref{alg:rmel}, and the same argument could be used for its variant Algorithm~\ref{alg:alterrmel}. Inspired by the techniques in~\cite{feng2022lipschitz}, we will show that under a clean process $\Phi$, the optimal arm $x_*$ would never be eliminated from layers whose tolerance levels are no less than $C$. Obviously, the optimal arm $x_*$ is in the covering when $m_{l,t} = 1$, where the whole arm space $\X$ is covered. Assume the layer $l_t$ reaches the end of epoch $m_{l_t,t}$ at time $t$ (i.e. $m_{l_t,t+1} = m_{l_t,t}+1$), and the optimal arm $x_*$ is contained in some active region $A_* \in A_{l_t,t}$. Then under a clean process, for any active region $A_0 \in A_{l_t,t}$ it holds that,
\begin{gather}
    f_{l_t,A_*,t} \geq \mu(x_*) - r_{l_t,A_*,t} \geq \mu(x_*) - {2}/{2^{m_{l_t,t}}} \label{eq:deltarmel1} \\
    f_{l_t,A_0,t} \leq \mu(x) + r_{l_t,A_0,t} \leq \mu(x) + {2}/{2^{m_{l_t,t}}}, \forall x \in A_0 \label{eq:deltarmel2}
\end{gather}
based on Lemma~\ref{lem:rmel4r} since we have $n_{l_t,A,t} = 6\ln(4T/\delta)\cdot 4^{m_{l_t,t}}, \forall A \in A_{l_t,t}$ in the end of the epoch. And since $\mu(x_*) \geq \mu(x),\forall x\in A_0$, it holds that
\begin{align}
    f_{l_t,A_0,t}- f_{l_t,A_*,t} \leq {4}/{2^{m_{l_t,t}}}. \label{eq:deltarmel3}
\end{align}
This implies that $A_*$ will not be removed. Note the above argument holds for any epoch index and any layer whose corruption level surpasses $C$, and hence the optimal arm $x_*$ would never be eliminated from layers whose tolerance levels are no less than $C$.

To prove Lemma~\ref{lem:deltarmel}. When $m_{l,t} = 1$, it naturally holds since $\Delta(x) \leq 1 \leq 16/2^{1}$. Otherwise, let $A_*$ be the covering that contains the optimal arm $x_*$ for layer $l$ in the previous epoch $m_{l,t}-1$, and according to the above argument it is well defined. And we know $x$ is also alive in the previous epoch, where we denote $A_x$ as the covering that contains $x$ in the previous epoch $m_{l,t}-1$. Denote $t_0$ as the time the last epoch reaches the end of layer $l$ ($m_{l,t}-1 = m_{l,t_0}$), and then it holds that
\begin{align*}
    \Delta (x) \leq f_{l,A_*,t_0}\! -\! f_{l,A_x,t_0} +2 r_{l,A_*,t_0} =  f_{l,A_*,t_0}\! -\! f_{l,A_x,t_0} +\frac{4}{2^{m_{l,t_0}}} = f_{l,A_*,t_0}\! - \!f_{l,A_x,t_0} +\frac{8}{2^{m_{l,t}}}
\end{align*}
since $r_{l,A_*,t_0} = r_{l,A_x,t_0} = {4}/{2^{m_{l,t_0}}}$ at the end of the epoch $m_{l,t_0}$. On the other hand, since $A_x$ was not eliminated at the end of the epoch $m_{l,t_0}$, based on the same argument used with Eqn. \eqref{eq:deltarmel1}, \eqref{eq:deltarmel2}, \eqref{eq:deltarmel3}, we have that
$$f_{l,A_*,t_0} - f_{l,A_x,t_0} \leq \frac{4}{2^{m_{l,t_0}}} = \frac{8}{2^{m_{l,t}}},$$
and this fact indicates that
$$\Delta (x) \leq \frac{16}{2^{m_{l,t}}}.$$
Note this result holds for any layer whose tolerance level surpasses $C$ and any $t \in [T]$. This implies Lemma~\ref{lem:deltarmel} holds conclusively.
\hfill \qedsymbol
\subsubsection{Proof of Theorem \ref{thm:rmel}}
\proof 
If the corruption budget $C \leq \ln{(4T/\delta)}$, then all the layers' tolerance levels exceed the unknown $C$, in which case based on Lemma~\ref{lem:cleanrmel}, with probability at least $1-3\delta /4$, it holds that $\forall x \in A, \forall A \in \A_{l,t}, \, \forall l \in [l^*], \, \forall t \in [T]$
\begin{align*}
    \left\lvert f_{l,A,t} - \mu(x) \right\rvert \leq\frac{1}{2^{m_l}} + & \sqrt{\frac{4 \ln{(T)} + 2 \ln{(4/\delta)}}{n_{l,A,t}}} + \frac{\ln{(T)} + \ln{(4/\delta)}}{n_{l,A,t}}.
\end{align*}
We denote $Regret_T(l)$ as the cumulative regret encountered from the layer $l$ across time $T$, which implies that
$$Regret_T = \sum_{l=1}^{l^*} Regret_T(l).$$
For any fixed layer $l \in [l^*]$, we will then show that $Regret_T(l) = \tilde O (T^{\frac{\zd+1}{\zd+2}})$. Based on Lemma~\ref{lem:deltarmel}, we know that for any layer $l$, any arm played after the epoch $m$ would at most incurs a regret of volume $16/2^m$. Note at epoch $m$, the active arm set consists of $1/2^m$-coverings for some region in $\{x \in \X: \Delta(x) \leq 16/2^m\}$. Therefore, the number of active regions at this epoch $m$ could be upper bounded
by $\alpha 2^{\zd m}$ for some constant $\alpha > 0$. And for each active region, we will pull it for exactly $6\ln(4T/\delta)\cdot 4^{m}$ times in epoch $m$. Therefore, the total regret incurred in the epoch $m$ for any layer would at most be
$$\alpha 2^{\zd m} \times 6\ln(4T/\delta)\cdot 4^{m} \times 16/2^m = 192 \alpha \ln(4T/\delta) 2^{(\zd+1)m}.$$
Therefore, the total cumulative regret we experience for any layer $l$ could be upper bounded as:
\begin{align*}
    Regret_T(l) &\leq \sum_{m=1}^M 192 \alpha \ln\left(\frac{4T}{\delta}\right) 2^{(\zd+1)m} + \frac{8}{2^M} T  \\
    &\leq 192 \alpha \ln\left(\frac{4T}{\delta}\right) \frac{2^{(\zd+1)(M+1)}-2^{\zd+1}}{2^{\zd+1} - 1} + \frac{8}{2^M} T \\
    &\leq 384 \alpha \ln\left(\frac{4T}{\delta}\right) 2^{(\zd+1)M} + \frac{8}{2^M} T, 
\end{align*}
where the second term bound the total regret after finishing the epoch $M$. Note we could take $M$ as any integer here, even if the epoch $M$ doesn't exist, our bound still works. By taking $M$ as the closest integer to the value $\left({\ln\left(\frac{T}{48 \alpha \ln(4T/\delta)} \right)}/\left[(\zd+2) \ln{(2)} \right] \right)$. It holds that
$$Regret_T(l) \lesssim T^{\frac{\zd+1}{\zd+2}} \ln{(T/\delta)}^{\frac{1}{\zd+2}}, \qquad \forall l \in [l^*].$$
Therefore, it holds that with probability at least $1-3\delta /4 \geq 1 - \delta$, 
$$Regret_T = \sum_{l=1}^{l^*} Regret_T(l) \lesssim T^{\frac{\zd+1}{\zd+2}} \ln{(T/\delta)}^{\frac{1}{\zd+2}} \cdot \log_2(T) = \tilde O (T^{\frac{\zd+1}{\zd+2}}).$$
On the other hand, if the corruption budget $C > \ln{(4T/\delta)}$, then not all the layers could tolerate the unknown total budget level $C$. We denote $l_0$ as the minimum index of the layer that is resilient to $C$ as defined in Eqn.~\ref{eq:phiset}. Therefore, we could use the above argument to similarly deduce that:
\begin{align}
\sum_{l=l_0}^{l^*} Regret_T(l) \lesssim T^{\frac{\zd+1}{\zd+2}} \ln{(T/\delta)}^{\frac{1}{\zd+2}} \cdot \log_2(T) = \tilde O (T^{\frac{\zd+1}{\zd+2}}). \label{eq:rmelfinal1}
\end{align}
For the first $(l_0 - 1)$ layers that are vulnerable to attacks, we could control their regret by using the cross-layer region elimination idea. Specifically, it holds that $v_{l_0} \leq BC$, then based on Corollary~\ref{coro:pull}, we know that with probability at least $1-\delta/4$, we select one time of layer $l_0$ at most every $BC \log(4T/\delta)$ times of the first $(l_0 - 1)$ non-robust layers. Since the active regions in a lower-index layer are always a subset of the active regions for the layer with a higher index according to our cross-layer elimination rule in Algorithm~\ref{alg:rmel}. We know when the layer $l_0$ stays at the epoch $m$, any arm played in the layer $1,2,\dots,l_0$ would at most incurs a regret $16/2^m$. Therefore, when the layer $l_0$ stays in epoch $m$, we have probability at least $1-3\delta/4-\delta/4 = 1- \delta$, the total regret incurred from the first $l_0$ layers altogether could be bounded as
$$BC \log(4T/\delta) \times \alpha 2^{\zd m} \times 6\ln(4T/\delta)\cdot 4^{m} \times 16/2^m = 192BC \alpha \ln(4T/\delta)^2 2^{(\zd+1)m}.$$
Conclusively, it holds that
\begin{align*}
    \sum_{l=1}^{l_0} Regret_T(l) &\leq \sum_{m=1}^M 192 BC \alpha \ln\left(\frac{4T}{\delta}\right)^2 2^{(\zd+1)m} + \frac{8}{2^M} T  \\
    &\leq 192 BC \alpha \ln\left(\frac{4T}{\delta}\right)^2 \frac{2^{(\zd+1)(M+1)}-2^{\zd+1}}{2^{\zd+1} - 1} + \frac{8}{2^M} T \\
    &\leq 384 BC \alpha \ln\left(\frac{4T}{\delta}\right)^2 2^{(\zd+1)M} + \frac{8}{2^M} T, 
\end{align*}
for arbitrary $M$. Similarly, then we can simply take $M$ as the closest positive integer to the value $$\left({\ln\left(\frac{T}{48 \alpha BC \ln(4T/\delta)} \right)}/\left[(\zd+2) \ln{(2)} \right] \right),$$ 
and we have that
\begin{align}
    \sum_{l=1}^{l_0} Regret_T(l) \lesssim T^{\frac{\zd+1}{\zd+2}} \left(BC \ln{(T/\delta)}^2 \right)^{\frac{1}{\zd+2}}. \label{eq:rmelfinal2}
\end{align}
Combine the results from Eqn.~\ref{eq:rmelfinal1} and Eqn.~\ref{eq:rmelfinal2}, with probability at least $1-\delta$, it holds that
$$Regret_T = \tilde O \left(T^{\frac{\zd+1}{\zd+2}} \left(B^{\frac{1}{\zd+2}}C^{\frac{1}{\zd+2}} +1 \right)  \right) = \tilde O \left(T^{\frac{\zd+1}{\zd+2}} \left(C^{\frac{1}{\zd+2}} +1 \right)  \right).$$
And this completes our proof.
\hfill \qedsymbol

%% file: app_thmmodelselection.tex
\subsection{Analysis of Theorem \ref{thm:model_selection}}\label{app:thmmodelselection}
\subsubsection{Useful Lemmas}
\begin{lemma}\label{lem:modelselection}
\textup{(Part of Theorem 3.2 and 5.3 in~\cite{pacchiano2020model})} If the regret of the optimal base algorithm could be bounded by $U_*(T,\delta) = O(c(\delta) T^\alpha)$ for some function $c:\mathbb{R} \rightarrow \mathbb{R}$ and constant $\alpha \in [1/2,1)$, the regret of EXP.P and CORRAL with smoothing transformation as the master algorithms are shown in Table~\ref{tab:lem}:
\begin{table}[ht]
\begin{center}
\caption{Table for Lemma~\ref{lem:modelselection}}\label{tab:lem}
\begin{tabular}{|c|c|}
\hline
 &Known $\alpha$, Unknown $c(\delta)$  \\\hline
EXP3.P  & $\tilde O\left(T^{\frac{1}{2-\alpha}}c(\delta)\right)$ \\\hline
CORRAL  & $\tilde O\left(T^{\alpha}c(\delta)^{\frac{1}{\alpha}}\right)$ \\\hline
\end{tabular}
\end{center}
\end{table}
\end{lemma}

The proof of this Lemma involves lots of technical details and is presented in~\cite{pacchiano2020model} elaborately. And hence we would omit the proof here.
\subsubsection{Proof of Theorem \ref{thm:model_selection}}
\proof 
The proof of our Theorem~\ref{thm:model_selection} is based on the above Lemma~\ref{lem:modelselection}. According to Theorem~\ref{thm:robust}, with probability at least $1/\delta$, the regret bound of our Algorithm~\ref{alg:robust} could be bounded as
$$Regret_T = \tilde O\left( T^{\frac{\zd+1}{\zd+2}} + C^{\frac{1}{\zd+1}} T^{\frac{\zd}{\zd+1}} \right) = \tilde O\left( T^{\frac{\zd+1}{\zd+2}} + C^{\frac{1}{\zd+2}} T^{\frac{\zd+1}{\zd+2}} \right).$$
Due to the fact that $\zd$ is upper bounded by $\cd$ and $C = O(T)$, it further holds that
$$Regret_T = \tilde O\left( \left(C^{\frac{1}{\zd+2}} +1 \right)T^{\frac{\zd+1}{\zd+2}} \right) = \tilde O\left( \left(C^{\frac{1}{\cd+2}} +1 \right)T^{\frac{\cd+1}{\cd+2}} \right).$$
Therefore, by taking $\alpha = \frac{\cd+1}{\cd+2}$ (known) and $c(\delta) = \left(C^{\frac{1}{\cd+2}} +1 \right)$ (unknown) and plugging them into Lemma~\ref{lem:modelselection}, we have that 
$$\mathbb{E}(Regret_T) = \begin{cases}
\tilde O\left((C^{\frac{1}{\cd+2}}+1) T^{\frac{\cd+2}{\cd+3}}\right) \quad \; \; \text{EXP3.P},\\
\tilde O\left((C^{\frac{1}{\cd+1}}+1) T^{\frac{\cd+1}{\cd+2}}\right)  \quad \; \; \text{CORRAL}.
\end{cases}$$
And this concludes our proof.
\hfill \qedsymbol

%% file: app_thmbob.tex
\subsection{Analysis of Theorem \ref{thm:bob}}\label{app:thmbob}
Under the assumption that the diameter of $\X$ is at most $1$, we could also assume that $\mu(x) \in [0,1], \forall x \in X$ due to the Lipschitzness of $\mu(\cdot)$ w.l.o.g. in this section.

\subsubsection{Useful Lemmas}
\begin{lemma}\label{lem:bound}
In Algorithm~\ref{alg:bob}, for any batch $i \in \left[\left\lceil\frac{T}{H}\right\rceil \right]$, the sum of stochastic rewards could be bounded as
$$\left | \sum_{t = (i-1)H+1}^{\min\{iH,T\}} y_t \right | \leq 2H + \sqrt{2H \log(\frac{12T}{H \delta})}$$
simultaneously with probability at least $1-\delta/3$.
\end{lemma}
\proof 
For arbitrary batch index $i \in \left[\left\lceil\frac{T}{H}\right\rceil \right]$, it holds that
\begin{align*}
    &P\left(\left | \sum_{t = (i-1)H+1}^{\min\{iH,T\}} y_t \right | \geq  2H + \sqrt{2H \log(\frac{12T}{H \delta})}\right)  \\
    = &P\left(\left | \sum_{t = (i-1)H+1}^{\min\{iH,T\}} \mu(x_t)+c_t(x_t) + \eta_t \right | \geq  2H + \sqrt{2H \log(\frac{12T}{H \delta})}\right) \\
    \leq&  P\left( \sum_{t = (i-1)H+1}^{\min\{iH,T\}} |\mu(x_t)|+ \sum_{t = (i-1)H+1}^{\min\{iH,T\}} |c_t(x_t)| + \left | \sum_{t = (i-1)H+1}^{\min\{iH,T\}} \eta_t  \right | \geq  2H + \sqrt{2H \log(\frac{12T}{H \delta})} \right)  \\
    \leq& P\left(  \left | \sum_{t = (i-1)H+1}^{\min\{iH,T\}} \eta_t  \right | \geq   \sqrt{2H \log(\frac{12T}{H \delta})} \right) \stackrel{\textup{(i)}}{\leq} \frac{H}{6T} \delta.
\end{align*}
The inequality (i) comes from the fact that $\sum_{t = (i-1)H+1}^{iH} \eta_t$ is sub-Gaussian with parameter $H$. Therefore, by taking a union bound on all $\left\lceil\frac{T}{H}\right\rceil$ batches, it holds that 
$$P\left(\forall i \in \left\lceil\frac{T}{H}\right\rceil : \left | \sum_{t = (i-1)H+1}^{\min\{iH,T\}} y_t \right | \leq  2H + \sqrt{2H \log(\frac{12T}{H \delta})}\right) \geq 1-\frac{\delta}{3}.$$
And this concludes the proof of Lemma~\ref{lem:bound}.\hfill\qedsymbol
\subsubsection{Proof of Theorem \ref{thm:bob}}
\proof
We are ready to prove Theorem \ref{thm:bob} now. Since we have $\lceil \log_2(T) \rceil$ base algorithms where the $i$-th base algorithm is our Robust Zooming Algorithm (Algorithm~\ref{alg:robust}) with tolerance level $2^i$, we can denote the base algorithm set as $W = \{ 2^i \}_{i=1}^{\lceil \log_2(T) \rceil}$ in terms of their tolerance levels. For any round $t \in [T]$, let $w_t$ denote the base algorithm chosen from $W$. And denote $x_t(w), \, w \in W$ as the arm pulled if the base algorithm $w$ is chosen in the beginning of its batch. In other words, we have $x_t = x_t(w_t)$. Denote $C_i$ as the total budget of corruptions in the $i$-th batch and hence $C = \sum_{i=1}^{\left\lceil{T}/{H}\right\rceil} C_i$, where recall that $C$ is the unknown total budget of corruptions. And we also write $C_* = \max_{i} C_i$ as the maximum budget in a single batch. Let $w_*$ be the element in $W$ such that $C_* \leq w_* < 2C_*$. Therefore, we could decompose the cumulative regret into the following two quantities:
 \begin{align}
    Regret_T  = \underbrace{ \sum_{t=1}^{T} \left( \mu(x_*) - \mu(x_t(w_*))\right)}_{\textstyle
    \begin{gathered}
       \text{Quantity (I)}
    \end{gathered}} + 
    \underbrace{\sum_{t=1}^{T} \left(  \mu(x_t(w_*)) -  \mu(x_t(w_t))\right)}_{\textstyle
    \begin{gathered}
       \text{Quantity (II)}
    \end{gathered}}  \label{eq:bob_divide}
\end{align}
And it suffices to bound these two quantities respectively. We know the Quantity (I) could be further represented as
$$\sum_{t=1}^{T} \left( \mu(x_*) - \mu(x_t(w_*))\right) = \sum_{i=1}^{\left\lceil \frac{T}{H} \right\rceil} \sum_{t=(i-1)H+1}^{\min\{iH,T\}} \left( \mu(x_*) - \mu(x_t(w_*))\right).$$
Here we will use the results from Theorem~\ref{thm:robust}. Note by setting the probability rate as $\delta/3$ in Algorithm~\ref{alg:robust}, we can prove that we have a clean process with probability at least $1-\delta/3$ (line 5 in Algorithm~\ref{alg:bob}). Although we run the Algorithm~\ref{alg:robust} here in a batch fashion and the total rounds is $T$, we can still easily show that with probability at least $1-\delta/3$ we have a clean process for all batches. This is because the proof of Lemma~\ref{lem:clean} only relies on taking a union bound over all rounds $T$ where whether a restart is proceeded doesn't matter at all. According to Theorem~\ref{thm:robust} and the choice of $w_*$, the cumulative regret of each batch could be upper bounded by the order of 
$$\tilde O \left(H^{\frac{\zd+1}{\zd+2}} + C_*^{\frac{1}{\zd+1}} H^{\frac{\zd}{\zd+1}}\right) = \tilde O \left(H^{\frac{\zd+1}{\zd+2}} + C_*^{\frac{1}{\cd+1}} H^{\frac{\cd}{\cd+1}}\right),$$
since $C_* \leq H$ naturally holds by definition. Therefore, it holds that
\begin{align}\label{eq:quantity1}
\text{Quantity (I)} = \tilde O \left( \left\lceil \frac{T}{H} \right\rceil \left( H^{\frac{\zd+1}{\zd+2}} + C_*^{\frac{1}{\cd+1}} H^{\frac{\cd}{\cd+1}}\right)\right) = \tilde O \left( TH^{\frac{-1}{\zd+2}} + TC_*^{\frac{1}{\cd+1}}  H^{\frac{-1}{\cd+1}} \right),
\end{align}
with probability at least $1-\delta/3$. For Quantity (II), according to Lemma~\ref{lem:bound}, for any batch $i \in \left[\left\lceil\frac{T}{H}\right\rceil \right]$ the sum of stochastic rewards could be bounded by
$$\left | \sum_{t = (i-1)H+1}^{\min\{iH,T\}} y_t \right | \leq 2H + \sqrt{2H \log\left(\frac{12T}{H \delta}\right)}$$
simultaneously with probability at least $1-\delta/3$. We denote the event $\Omega$ as
$$\Omega = \left\{ \forall i \in \left\lceil\frac{T}{H}\right\rceil : \left | \sum_{t = (i-1)H+1}^{\min\{iH,T\}} y_t \right | \leq  2H + \sqrt{2H \log\left(\frac{12T}{H \delta}\right)} \right\},$$
and it holds that $P(\Omega) \geq 1 - \delta/3$.
And under the event $\Omega$, from Theorem 6.3 in~\cite{auer2002nonstochastic}, we know with probability at least $1-\delta/3$, it holds that 
\begin{align}\label{eq:quantity2}
\text{Quantity (II)} = \tilde O \left( \sqrt{H^2 \frac{T}{H}} \right) = \tilde O \left( \sqrt{TH} \right).
\end{align}
Specifically, in the statement of Theorem 6.3~\cite{auer2002nonstochastic}, we have $K = \lceil \log_2(T) \rceil, \delta = \delta/3, T = \lceil \frac{T}{H} \rceil$ here. And we multiply the regret bound in Theorem 6.3~\cite{auer2002nonstochastic} by $\left(2H + \sqrt{2H \log(\frac{12T}{H \delta})}\right)$ as well since the original EXP3.P algorithm requires the magnitude of rewards not exceeding $1$. Conclusively, by combining the results from Eqn.~\ref{eq:quantity1} and Eqn.~\ref{eq:quantity2} and taking a union bound on the probability rates, with probability at least $1-\delta/3-\delta/3-\delta/3 = 1 - \delta$, we have that 
$$Regret_T = \tilde O \left(TH^{\frac{-1}{\zd+2}} + TC_*^{\frac{1}{\cd+1}}  H^{\frac{-1}{\cd+1}} + \sqrt{TH} \right).$$
By taking $H = T^\frac{\cd+2}{\cd+4}$ and using the fact that $\zd \leq \cd$, it holds that
\begin{align*}
Regret_T = \tilde O \left( T^\frac{\cd+3}{\cd+4} + C_*^\frac{1}{\cd+1} T^\frac{\cd^2+4\cd+2}{(\cd+1)(\cd+4)} \right) &= \tilde O \left( T^\frac{\cd+3}{\cd+4} + C_*^\frac{1}{\cd+1} T^\frac{\cd+2}{\cd+3} \right)  
\\ &= \tilde O \left( T^\frac{\cd+3}{\cd+4} + C^\frac{1}{\cd+1} T^\frac{\cd+2}{\cd+3} \right), 
\end{align*}
with probability at least $1-\delta$.
\hfill \qedsymbol

%% file: app_alteralg.tex
\subsection{Additional Algorithms}\label{app:alteralg}
\subsubsection{Alternative Algorithm for RMEL}

Here we present another version of the RMEL algorithm in~\ref{alg:alterrmel}. Instead of executing the elimination process after each epoch of any layer as in~\ref{alg:rmel}, here we conduct the elimination at each round. This modification will make our algorithm more accurate and discard less promising regions in a timely manner but will lead to higher computational complexity as well. Note the regret bound of Theorem \ref{thm:rmel} naturally holds since we could use the identical proof to reach the same regret bound here. And we also add an explanation before Corollary~\ref{coro:pull} in Appendix~\ref{app:thmrmel}.

\begin{algorithm}[t]
\caption{Alternative Robust Multi-layer Elimination Lipschitz Bandit Algorithm (RMEL)} \label{alg:alterrmel}
\begin{algorithmic}[1]
\Input Arm metric space $(\X,D)$, time horizon $T$, probability rate $\delta$, base parameter $B$.
\Stage Tolerance level $v_l = \ln{(4T/\delta)} B^{l-1}, m_l=1, n_l = 0, \A_l = 1/2$-covering of $\X, \, f_{l,A} = n_{l,A} = 0$ for all $A \in \A_l, l \in [l^*]$ where $l^* \coloneqq \min \{l \in \mathbb{N} : \ln{(4T/\delta)} B^{l-1} \geq T \}$.
\For{$t=1$ {\bfseries to} $T$}
\State Sample layer $l \in [l^*]$ with probability $1/v_l$, with the remaining probability sampling $l=1$. Find the minimum layer index $l_t \geq l$ such that $\A_{l_t} \neq \emptyset$. {\color{red} \Comment{Layer sampling}}
\State Choose $A_t = \arg\min_{A \in \A_{l_t}} n_{l_t,A}$, break ties arbitrary.
\State Randomly pull an arm $x_t \in A_t$, and observe the payoff $y_t$.
\State Set $n_{l_t} = n_{l_t}+1, \, n_{l_t,A_t}= n_{l_t,A_t}+1,\text{and } f_{l_t,A_t} =\left(f_{l_t,A_t} (n_{l_t,A_t}-1) + y_t\right)/n_{l_t,A_t}$.
\State {Obtain $f_{l_t,*} = \max_{A \in \A_{l_t}} f_{l_t,A}, \, n_{l_t,*} = \min_{A \in \A_{l_t}} n_{l_t,A}$.}
\State {For each $A \in \A_{l_t}$, if $f_{l_t,*} -  f_{l_t,A} > 2/2^{m_{l_t}}+\sqrt{8\ln{(4T^2/\delta)}/n_{l_t,*}}+2\ln{(4T/\delta)}/n_{l_t,*}$, then we eliminate $A$ from $\A_{l_t}$ and all active regions $A'$ from $\A_{l'}$ in the case that $A' \subseteq A, A' \in \A_{l'}, l' < l$. }{\color{red}\Comment{Removal}}
\If{$n_{l_t} = 6\ln(4T/\delta)\cdot 4^{m_l} \times |\A_{l_t}|$} 
\State Find $1/2^{m_l+1}$-covering of each $A \in \A_{l_t}$ in the same way as $A$ was partitioned in other layers. Then reload the active region set $\A_{l_t}$ as the collection of these coverings.
\State Set $n_{l_t} = 0, \, m_{l_t} = m_{l_t}+1$. And renew $n_{l_t,A} = f_{l_t,A} = 0, \forall A \in \A_{l_t}$. {\color{red}\Comment{Refresh}}
\EndIf
\EndFor
\end{algorithmic}
\end{algorithm}

\subsubsection{BoB Robust Zooming Algorithm}
Due to the space limit, we defer the pseudocode of BoB Robust Zooming algorithm here in Algorithm \ref{alg:bob}. We can observe that the top layer is an EXP3.P algorithm, which chooses the corruption level used for Robust Zooming algorithm in each batch adaptively. For each batch, we run our Robust Zooming algorithm with the chosen corruption level from the top layer, and use the accumulative rewards collected in each batch to update the components of EXP3.P (i.e. line 10 of Algorithm \ref{alg:bob}). Note we normalize the cumulative reward by dividing it with $(2H + \sqrt{2H \log(\frac{12T}{H \delta})})$, and this is because that we could prove that the magnitude of the cumulative reward at each batch would be at most $(2H + \sqrt{2H \log(\frac{12T}{H \delta})})$ with high probability as shown in Lemma~\ref{lem:bound}. And the EXP3.P algorithm~\cite{auer2002nonstochastic} requires the magnitude of reward should at most be $1$ with our chosen values of $\alpha$ and $\gamma$. The regret bound of Algorithm \ref{alg:bob} is given in Theorem \ref{thm:bob} of our main paper.

\begin{algorithm}[t]
\caption{BoB Robust Zooming Algorithm} \label{alg:bob}
\begin{algorithmic}[1]
\Input Arm metric space $(\X,D)$, time horizon $T$, probability rate $\delta$, batch size $H$.
\Stage Budget set for base algorithms $I = \{2^i\}_{i=1}^{N}, \, N =\lceil \log_2(T) \rceil $, $\alpha = 2 \sqrt{\ln{(\frac{3NT}{\delta})}}, \gamma  = \min \left\{\frac{3}{5}, 2 \sqrt{ \frac{3 N \ln{(N)}}{5T}} \right\}$, weight $w_i = 1, i \in [N]$, cumulative sum $s=0$.
\For{$t=1$ {\bfseries to} $T$}
\If{$t \in \{kH+1 : k \in \mathbb{N}\}$}
\State For $i = 1, \dots, N$ set
$$p_i = (1-\gamma) \frac{w_i}{\sum_{j=1}^N w_j} + \frac{\gamma}{N}.$$
\State Choose the base algorithm index $i'$ randomly with probability $\{p_i\}_{i=1}^N$.
\State Refresh the chosen Robust Zooming algorithm (Algorithm \ref{alg:robust} with $C = 2^{i'}$) with active arm set $J = \{\}$, active space $\X_{act} = \X$ and probability rate $\delta/3$.
\EndIf
\State Run the chosen Robust Zooming algorithm and receive the reward $y_t$.
\State Update the chosen Robust Zooming algorithm according to Algorithm \ref{alg:robust} and set $s = s + y_t$.
\If{$t \in \{kH : k \in \mathbb{N}^+\}$}
\State Let $s = {s}/\left[p_{i'} \left(2H + \sqrt{2H \log(\frac{12T}{H \delta})} \right)\right]$.
\State Update EXP3.P component for index $i'$: $w_{i'} = w_{i'} \exp \left( \frac{\gamma}{3N} \left(s + \frac{\alpha}{p_{i'} \sqrt{NT}}  \right)\right)$.
\EndIf
\EndFor
\end{algorithmic}
\end{algorithm}

%% file: app_lowerbound.tex
\subsection{Discussion on Lower Bounds}\label{app:lowerbound}

We now propose Theorem \ref{thm:lowerbounds} and Theorem \ref{thm:lowerboundw} with their detailed proof in Section~\ref{app:lowerbounds} and Section \ref{app:lowerboundw} respectively, where we provide a pair of lower bounds for the strong adversary and the weak adversary.

\subsubsection{Lower Bound for Strong Adversaries}\label{app:lowerbounds}
We repeat our Theorem \ref{thm:lowerbounds} for reference here and then provide a detailed proof as follows:

\textbf{Theorem \ref{thm:lowerbounds}} \textit{
Under the strong adversary with corruption budget $C$, for any zooming dimension $\dz \in \mathbb{Z}^+$, there exists an instance such that any algorithm (even is aware of $C$) must suffer from the regret of order $\Omega \left(C^{\frac{1}{\zd+1}}T^{\frac{\zd}{\zd+1}} \right)$ with probability at least $0.5$.}

\proof Here we consider the metric space $([0,1)^d,l_{\infty})$. For arbitrary $\epsilon \in (0,\frac{1}{2})$, we can equally divide the space $[0,1]^d$ into $1/\epsilon^d$ small $l_{\infty}$ balls whose diameters are equal to $\epsilon$ by discretizing each axis. (W.l.o.g we assume $1$ is divisible by $\epsilon$ for simplicity since otherwise we could take $\lfloor 1/\epsilon^d \rfloor$ instead.) For example, if $d=2$ and $\epsilon=\frac{1}{2}$, then we can divide the space into $2^2=4$ $l_{\infty}$ balls: $[0,0.5)^2, [0,0.5) \times [0.5,1), [0.5,1) \times [0,0.5), [0.5,1)^2$. We denote these balls as $\{A_i\}_{i=1}^{1/\epsilon^d}, [0,1)^d = \cup_{i=1}^{1/\epsilon^d} A_i$ and their centers as $\{c_i\}_{i=1}^{1/\epsilon^d}$. (e.g. the center of $[0,0.5)^2$ is $(0.25,0.25)$.) Subsequently, we could define a set of functions $\{f_i(\cdot)\}_{i=1}^{1/\epsilon^d}$ as
$$
f_i(x) = \begin{cases}
\frac{\epsilon}{2} - \|x - c_i \|_{\infty}, \quad & x \in A_i; \\
0, \quad &x \notin A_i.
\end{cases}$$
We can easily verify that $f_i(\cdot)$ is a $1$-Lipschitz function. For the zooming dimension, if $\epsilon$ is of constant scale, then the zooming dimension will become $0$. However, in our analysis here, we would let $\epsilon$ rely on $T$ and be sufficiently small so that the zooming dimension is $d$. If the underlying expected reward function is $f_k(\cdot)$ and there is no random noise, consider the strong adversary that shifts the reward of the arm down to whenever the pulled arm is in $A_k$ and doesn't attack the reward otherwise. This attack could be done for roughly $\lfloor C/\epsilon \rfloor$ times. Intuitively, the learner can do no better than pull each arm in $[0,1]^d$ uniformly. This implies that roughly the learner should do $\lfloor C/\epsilon \rfloor \lfloor 1/\epsilon^d \rfloor$ rounds of uniform exploration before the attack budget $C$ is used up, where the learner pulls arms outside $A_k$ for approximately $\lfloor C/\epsilon \rfloor \cdot \lfloor (1-\epsilon^d)/\epsilon^d \rfloor$ times. Take $\epsilon = \left( \frac{C}{T}\right)^{\frac{1}{d+1}}$, we know that roughly the learner should do $\lfloor C/\epsilon \rfloor \lfloor 1/\epsilon^d \rfloor = T$ rounds of uniform exploration, and the cumulative regret is at least
$$\left\lfloor \frac{C}{\epsilon} \right\rfloor \cdot \left\lfloor \frac{(1-\epsilon^d)}{\epsilon^d} \right\rfloor \cdot \epsilon =\Theta \left(C^{\frac{1}{d+1}}T^{\frac{d}{d+1}} \right)= \Theta \left(C^{\frac{1}{\zd+1}}T^{\frac{\zd}{\zd+1}} \right).$$

For a more rigorous argument, note that for the $k$-th instance $f_k(\cdot)$, the adversary could maliciously replace the reward with $0$ until the arm in $A_k$ is pulled at least $\lfloor C/\epsilon \rfloor$ times. After $\lfloor C/\epsilon \rfloor \lfloor 1/2\epsilon^d \rfloor $ rounds, for any algorithm even with the information of value $C$, there must be at least $\lfloor 1/(2\epsilon^d) \rfloor$ balls among $\{A_i\}_{i=1}^{1/\epsilon^d}$ that have been pulled for at most $\lfloor C/\epsilon \rfloor$ times. As a consequence, when we choose the problem instance $k$ among these $\lfloor 1/(2\epsilon^d) \rfloor$ balls and set $\epsilon = \left( \frac{C}{T}\right)^{\frac{1}{d+1}}$, then we know that the regret of order 
$$\epsilon \cdot \left\lfloor \frac{C}{\epsilon} \right\rfloor \cdot \left(  \left\lfloor \frac{1}{2\epsilon^d} \right\rfloor - 1 \right) = \Theta \left( C^{\frac{1}{\zd+1}}T^{\frac{\zd}{\zd+1}} \right)$$ 
is unavoidable. This implies that the regret could be no worse than $\Omega ( C^{\frac{1}{\zd+1}}T^{\frac{\zd}{\zd+1}} )$ under the strong adversary with probability $0.5$.
\hfill \qedsymbol

For the stochastic Lipschitz bandit problem, based on~\cite{slivkins2011contextual} we know for any algorithm there exists one problem instance such that the expected regret is at least
$$\inf_{r_0 \in (0,1)} \left( r_0 T + C \log(T) \sum_{r = 2^{-i}: i \in \mathbb{N}, r \geq r_0} \frac{N_z(r)}{r} \right),$$
where $N_z(r)$ is the zooming number. And hence the corruption-free lower bound $O \left(\ln(T)^{\frac{1}{\zd+2}} T^{\frac{\zd+1}{\zd+2}} \right)$ is optimal in terms of the zooming dimension $\zd$. Combining this result with our Theorem~\ref{thm:lowerbounds}, we can conclude that for any algorithm, there exists a corrupted bandit instance where the algorithm must incur $\Omega \left( \max\left\{\ln(T)^{\frac{1}{\zd+2}} T^{\frac{\zd+1}{\zd+2}}, C^{\frac{1}{\zd+1}}T^{\frac{\zd}{\zd+1}} \right\} \right)$ cumulative regret, which coincides with the order of regret for our Robust Zooming algorithm. Conclusively, our algorithm obtains the optimal order of regret under the strong adversary.

We then restate our Theorem \ref{thm:lowerboundsknown} for reference and then provide a detailed proof:

\textbf{Theorem \ref{thm:lowerboundsknown}} \textit{
For any algorithm, when there is no corruption, we denote $R_T^0$ as the upper bound of cumulative regret in $T$ rounds under our problem setting described in Section~\ref{sec:preliminaries}, i.e. $Regret_T \leq R_T^0$ with high probability, and it holds that $R_T^0 = o(T)$. Then under the strong adversary and unknown attacking budget $C$, there exists a problem instance on which this algorithm will incur linear regret $\Omega(T)$ with probability at least $0.5$, if $C = \Omega(R_T^0/4^{d_z})=\Omega(R_T^0)$.}
\proof
For the case that $d_z=0$, we consider the metric space $([0,1], l_2)$ and define the Lipschitz function $f_1(\cdot)$ as
\hfill \qedsymbol
$$
f_1(x) = \begin{cases}
0.25 - |x-0.25|, \quad & x \in [0,0.5]; \\
0, \quad &x \in (0.5,1].
\end{cases},$$
and we assume there is no random noise and no adversarial corruption. (We call this instance $I_0$.) For any algorithm with $\mathbb{E}(Regret_T) \leq R_T^0$ when there is no adversarial corruption, we know that 
$$\mathbb{E}(\#\text{ iterations playing arms in } (0.5,1]) \times 0.25 \leq \mathbb{E}(Regret_T) \leq R_T^0,$$
and hence $\mathbb{E}(\#\text{ iterations playing arms in } (0.5,1]) \leq 4R_T^0$. By Markov inequality, with probability at least $0.5$, the number of iterations that play arms in $(0.5,1]$ is no more than $8R_T^0$.

Next, we define a new problem setting in the same metric space as:
$$
f_2(x) = \begin{cases}
0.25 - |x-0.25|, \quad & x \in [0,0.5]; \\
x - 0.5, \quad &x \in (0.5,1].
\end{cases}.$$
And under the setting of $f_2(\cdot)$ there is a malicious strong adversary with budget $C = 4R_T^0$ to attack using the following strategy: whenever the arm in $(0.5,1]$ is selected and the corruption budget has not been used up, the adversary moves the reward to $0$. We call this instance $I_1$. Therefore, before the budget is used up, each selection of arm in $(0.5,1]$ returns a reward $0$, and hence the agent can never tell the difference between $I_0$ and $I_1$ and would follow the same strategy under $I_0$ until the total corruption level reaches $C = 4R_T^0$ and then the adversary stops to contaminate the rewards. And this requires at least $C/0.5 = 2C = 8R_T^0$ rounds in which the agent chooses arms in $(0.5,1]$. Therefore, with probability of at least $0.5$, the regret in $T$ rounds is at least $(T - 8R_T^0)/4 = \Omega(T)$.

For $d_z > 0$, we use the metric space $([0,1]^d, \|\cdot\|_{\infty})$ with $d = \lceil 2d_z \rceil$. We first partition the $d$-dimensional cube $[0,1]^d$ into $2^d$ sub-cubes with side length $0.5$, i.e. equally divide the cube $[0,1]^d$ into $0.5$-radius $l_{\infty}$ balls whose diameters are equal to $0.5$ by discretizing each axis. We denote these balls as ${A_i}_{i=1}^{2^d}$ and the center of these balls as ${c_i}_{i=1}^{2^d}$, e.g. $c_1 = [0.25]^d$. And we denote the vertex of each ball that matches the vertexes of $[0,1]^d$ as ${v_i}_{i=1}^{2^d}$, e.g. $v_1 = [0]^d$. Subsequently, we could define the function $f_1(\cdot)$ as
$$
f_1(x) = \begin{cases}
{4^{\frac{-d}{d-\zd}}} - \|x - c_1 \|_{\infty}^{\frac{d}{d-\zd}}, \quad & x \in A_1; \\
0, \quad &x \notin A_1.
\end{cases}$$
and we assume there is no random noise and no adversarial corruption. (We call this instance $I_0$.) Since the regret of the algorithm under no corruption satisfies that $\mathbb{E}(Regret_T) \leq R_T^0$, and we know that pulling any arm outside $A_1$ will incur a single regret of $4^{\frac{-d}{d-\zd}}$, and hence we have that
$$\mathbb{E}(\#\text{ iterations playing arms not in } A_1) \leq R_T^0 \cdot 4^{\frac{-d}{d-\zd}}.$$
Then by the pigeonhole principle, there exists a sub-ball $2\leq i \leq 2^d$ such that the expected number of iterations to pull arms in $A_i$ is no more than $R_T^0 \cdot 4^{\frac{-d}{d-\zd}}/(2^d-1)$. Without loss of generality, we assume $i=2$, where $c_2 = [0.75,0.25,\dots,0.25]$ and $v_2 = [1,0,\dots,0]$. Similarly by using Markov Inequality, with probability at least $0.5$, the number of iterations that play arms in $A_2$ is no more than $2R_T^0 \cdot 4^{\frac{-d}{d-\zd}}/(2^d-1)$.

Next, we define a new problem setting in the same metric space as:
$$
f_2(x) = \begin{cases}
{4^{\frac{-d}{d-\zd}}} - \|x - c_1 \|_{\infty}^{\frac{d}{d-\zd}}, \quad & x \in A_1; \\
{2^{\frac{-d}{d-\zd}}} - \|x - v_2 \|_{\infty}^{\frac{d}{d-\zd}}, \quad & x \in A_2; \\
0, \quad &x \notin A_1 \cup A_2.
\end{cases}.$$
And under the setting of $f_2(\cdot)$ there is a malicious strong adversary with budget $C = 2R_T^0 \cdot 2^{\frac{-d}{d-\zd}}/(2^d-1) = \Theta(R_T^0/2^d)$ to attack the rewards. (Note $1 \leq {d}/{(d-\zd)} \leq 2$). Specifically, the adversary uses the following strategy: whenever the arm in $A_2$ is selected and the corruption budget has not been used up, the adversary moves the reward to $0$. We call this instance $I_1$. Therefore, before the budget is used up, each selection of arm in $A_2$ returns a reward $0$, and hence the agent can never tell the difference between $I_0$ and $I_1$ and would follow the same strategy under $I_0$ until the total corruption level reaches $C = 2R_T^0 \cdot 2^{\frac{-d}{d-\zd}}/(2^d-1)$, and then the adversary stops to contaminate the rewards. And this requires at least $C/2^{\frac{-d}{d-\zd}} =  2R_T^0 \cdot 4^{\frac{-d}{d-\zd}}/(2^d-1)$ rounds in which the agent chooses arms in $A_2$. Therefore, with probability of at least $0.5$, the regret in $T$ rounds is at least $$\left(T - \frac{2 4^{\frac{-d}{d-\zd}} R_T^0}{2^d-1} \right) \times \left({2^{\frac{-d}{d-\zd}}} - {4^{\frac{-d}{d-\zd}}}\right) \geq \frac{3}{16} \left(T - \frac{32 R_T^0}{2^d-1} \right)  = \Omega(T).$$ \hfill\qedsymbol

\subsubsection{Lower Bound for Weak Adversaries}\label{app:lowerboundw}

Recall Theorem~\ref{thm:lowerboundw} in our main paper:

\textbf{Theorem~\ref{thm:lowerboundw}} \textit{Under the weak adversary with corruption budget $C$, for any zooming dimension $\dz$, there exists an instance such that any algorithm (even is aware of $C$) must suffer from the regret of order $\Omega (C)$ with probability at least $0.5$.}

\proof 
We can modify the argument of the previous subsection~\ref{app:lowerbounds} to validate Theorem~\ref{thm:lowerboundw}. If $\zd = 0$, we could simply use the metric space $([0,1), l_2)$ and the reward function
$$
\mu_1(x) = \begin{cases}
\frac{1}{2} - |x - \frac{1}{4}|, \quad & x \in [0,0.5); \\
0, \quad &x \in [0.5,1).
\end{cases} \quad \; \;
\mu_2(x) = \begin{cases}
0, \quad & x \in [0,0.5); \\
\frac{1}{2} - |x - \frac{3}{4}|, \quad &x \in [0.5,1).
\end{cases}$$
We can easily verify that the zooming dimension $\zd = 0$ holds. Assume there is no random noise, and at each iteration the weak adversary pushes the reward everywhere in $[0,1)$ to $0$, which would use a $0.5$ budget. Therefore, this attack could last for the first $\lfloor 2C \rfloor$ rounds, when the agent would just receive a $0$ reward regardless of the pulled arm. For any algorithm, it would at least spend for $\lfloor C \rfloor$ rounds on either $[0,0.5)$ or $[0.5,1)$ with probability at least $0.5$. By considering the above two reward functions, we know that it would incur $\Omega(C)$ regret with probability at least $0.5$.

For $\zd > 0$, we set $d = \lceil 2\zd \rceil$ and consider the metric space $([0,1)^d, l_\infty)$. Similarly, we can equally divide the space $[0,1]^d$ into $1/2$ small $l_{\infty}$ balls whose diameters are equal to $1/2$ by discretizing each axis. We denote these balls as $\{A_i\}_{i=1}^{2^d}, [0,1)^d = \cup_{i=1}^{2^d} A_i$ and their centers as $\{c_i\}_{i=1}^{2^d}$. (e.g. the center of $[0,0.5)^2$ is $(0.25,0.25)$.) Subsequently, we could define a set of functions $\{f_i(\cdot)\}_{i=1}^{1/2^d}$ as
$$
\mu_i(x) = \begin{cases}
{4^{\frac{-d}{d-\zd}}} - \|x - c_i \|_{\infty}^{\frac{d}{d-\zd}}, \quad & x \in A_i; \\
0, \quad &x \notin A_i.
\end{cases}$$
We can easily verify that the zooming dimension of any instance is $\zd$. Assume there is no random noise, and at each iteration, the weak adversary pushes the reward everywhere in $[0,1)$ to $0$, which would use a $4^{\frac{-d}{d-\zd}}$ budget. Therefore, this attack could last for the first $\lfloor 4^{\frac{d}{d-\zd}}C \rfloor$ rounds, when the agent would just receive a $0$ reward regardless of the pulled arm. After $\lfloor 4^{\frac{\zd}{d-\zd}}C \rfloor$ rounds, for any algorithm even with the information of value $C$, there must be at least $\lfloor 2^{d-1} \rfloor$ balls among $\{A_i\}_{i=1}^{2^d}$ that have been pulled for at most $\lfloor 2^{(\frac{2d}{d-\zd} - d)} C \rfloor = \Theta(C)$ times. As a consequence, as for the problem instance $k$ among these $\lfloor 2^{d-1} \rfloor$ balls, the regret incurred $\Omega(C)$. Similarly, this means that any algorithm must incur $\Omega(C)$ regret with probability $0.5$.
\hfill\qedsymbol

%% file: app_exp.tex
\subsection{Additional Experimental Details}\label{app:exp}

Note in our main paper we assume that $\sigma = 1$, and our pseudocodes of Algorithms are based on this assumption. When we know a better upper bound for $\sigma$, we could easily modify the components in each algorithm based on $\sigma$. For example, we could modify the confidence radius of any active arm $x$ in Algorithm \ref{alg:robust} as
$$r(x) = \sigma\sqrt{\frac{4 \ln{(T)} + 2 \ln{(2/\delta)}}{n(x)}} + \frac{C}{n(x)}.$$

Next, we exhibit the setup of algorithms involved in our experiments as follows:
\begin{itemize}
    \item \textbf{Zooming algorithm~\cite{kleinberg2019bandits}:} We use the same setting for stochastic Lipschitz bandit as in~\cite{kleinberg2019bandits}, and set the radius for each arm as:
    $$r(x) = \sigma\sqrt{\frac{4 \ln{(T) + 2 \ln{(2/\delta)}}}{n(x)}}.$$
    And its implementation is available with the library~\cite{Li2023PyXAB}.
    \item \textbf{RMEL (ours):} We use the same parameter setting for RMEL as shown in our Algorithm \ref{alg:rmel}. And based on the experimental results in Figure~\ref{plt:exp}, this method apparently works best under different kinds of attacks and reward functions. 
    \item \textbf{BoB Robust Zooming algorithm (ours):} We use the same parameter setting with $\sigma$ for BoB Robust Zooming algorithm as shown in Algorithm~\ref{alg:bob} without restarting the algorithm after each batch since we found that restarting will sometimes abandon useful information empirically. 
    This BoB-based approach also works well according to Figure \ref{plt:exp}.
\end{itemize}
The numerical results of final cumulative regrets in our simulations in Section~\ref{sec:exp} (Figure~\ref{plt:exp}) are displayed in Table~\ref{tab:exp}. 

Note our RMEL (Algorithm~\ref{alg:rmel}) is designed to defend against the weak adversary in the theoretical analysis, and hence to be consistent, we also consider the weak adversary for both types of attacks under the same experimental setting and three levels of corrupted budgets. Recall that in the previous experiments in Section~\ref{sec:exp}, the adversary will contaminate the stochastic rewards only if the pulled action is in the specific region (Oracle: benign arm, Garcelon: targeted arm region), and otherwise the adversary will not spend its budget. And hence it is a strong adversary whose action relies on the current arm. 
To adapt these two attacks into a weak-adversary version, we could simply inject both sorts of attacks at each round based on their principles at each round: the Oracle will uniformly push the expected rewards of all ``good arms'' below the expected reward of the worst arm with an additional margin of $0.1$ with probability $0.5$ at the very beginning of each round. And the Garcelon will modify the expected rewards of all arms outside the targeted region into a random Gaussian noise $N(0,0.01)$ with probability $0.5$ ahead of the agent's action. Consequently, adversaries may consume the corruption budget at each round regardless of the pulled arm, and we expect that they will run out of their total budget in fewer iterations than the strong adversary does. We use the same experimental settings as in Section~\ref{sec:exp}, and the results are exhibited in Table~\ref{tab:exp2}.

From Table~\ref{tab:exp2}, we can see the experimental results under the weak adversary are consistent with those under the strong adversary. The state-of-the-art Zooming algorithm is evidently vulnerable to the corruptions, while our proposed algorithms, especially RMEL, could yield robust performance across multiple settings consistently. We can also observe that compared with the strong adversary, the weak adversary is less malicious than expected.

Another remark is that the adversarial settings used in our experiments may not be consistent with the assumption that $|c_t(x)| \leq 1$, while we find that (1). by modifying the original attacks and restricting the attack volume to be at most one with truncation, we can get a very similar result as shown in Table~\ref{tab:exp} and Table~\ref{tab:exp2}. (2). actually we can change the assumption to $|c_t(x)| \leq u$ where $u$ is an arbitrary positive constant for the theoretical analysis.

\begin{table}[ht]
\begin{center}
\begin{tabular}{lcc|cc}
\hline \hline
\multicolumn{1}{l}{\multirow{10}{*}{Triangle reward function}} &  \multicolumn{1}{|c|}{Algorithm}                           & Budget ($C$) & Oracle   & Garcelon \\ \cline{2-5}
&\multicolumn{1}{|c|}{\multirow{3}{*}{Zooming}}            & 0            & 366.58   & 366.58   \\
&\multicolumn{1}{|c|}{}                                    & 3000         & 10883.51 & 10660.17 \\
&\multicolumn{1}{|c|}{}                                    & 4500         & 11153.78 & 11487.59 \\\cline{2-5}
&\multicolumn{1}{|c|}{\multirow{3}{*}{RMEL}}               & 0            & 512.46   & 512.46   \\
&\multicolumn{1}{|c|}{}                                    & 3000         & 921.95   & 504.78   \\
&\multicolumn{1}{|c|}{}                                    & 4500         & 928.27   & 1542.17  \\\cline{2-5}
&\multicolumn{1}{|c|}{\multirow{3}{*}{BoB Robust Zooming}} & 0            & 461.16   & 461.16   \\
&\multicolumn{1}{|c|}{}                                    & 3000         & 495.06   & 531.37   \\
&\multicolumn{1}{|c|}{}                                    & 4500         & 1323.97  & 736.85   \\ \hline \hline
\multicolumn{1}{l}{\multirow{10}{*}{Sine reward function}} & 
\multicolumn{1}{|c|}{Algorithm}                           & Budget ($C$) & Oracle  & Garcelon \\\cline{2-5}
&\multicolumn{1}{|c|}{\multirow{3}{*}{Zooming}}            & 0            & 315.94  & 315.94   \\
&\multicolumn{1}{|l|}{}                                    & 3000         & 5289.65 & 3174.26  \\
&\multicolumn{1}{|l|}{}                                    & 4500         & 5720.30 & 3174.29  \\ \cline{2-5}
&\multicolumn{1}{|c|}{\multirow{3}{*}{RMEL}}               & 0            & 289.86  & 289.86   \\
&\multicolumn{1}{|l|}{}                                    & 3000         & 442.66  & 289.29   \\
&\multicolumn{1}{|l|}{}                                    & 4500         & 862.90  & 332.71   \\ \cline{2-5}
&\multicolumn{1}{|c|}{\multirow{3}{*}{BoB Robust Zooming}} & 0            & 435.44  & 435.44   \\
&\multicolumn{1}{|c|}{}                                    & 3000         & 414.54  & 746.96   \\
&\multicolumn{1}{|c|}{}                                    & 4500         & 1887.35 & 1148.09  \\ \hline \hline
\multicolumn{1}{l}{\multirow{10}{*}{Two dim reward function}} & 
\multicolumn{1}{|c|}{Algorithm}                           & Budget ($C$) & Oracle  & Garcelon \\ \cline{2-5}
&\multicolumn{1}{|c|}{\multirow{3}{*}{Zooming}}            & 0            & 3248.54 & 3248.54  \\
&\multicolumn{1}{|l|}{}                                    & 3000         & 8730.73 & 8149.79  \\
&\multicolumn{1}{|l|}{}                                    & 4500         & 9496.83 & 13672.00 \\ \cline{2-5}
&\multicolumn{1}{|c|}{\multirow{3}{*}{RMEL}}               & 0            & 2589.32 & 2589.32  \\
&\multicolumn{1}{|l|}{}                                    & 3000         & 5660.10 & 2590.77  \\
&\multicolumn{1}{|l|}{}                                    & 4500         & 6265.09 & 2872.64  \\ \cline{2-5}
&\multicolumn{1}{|c|}{\multirow{3}{*}{BoB Robust Zooming}} & 0            & 3831.94 & 3831.94  \\
&\multicolumn{1}{|c|}{}                                    & 3000         & 6310.29 & 4217.74  \\
&\multicolumn{1}{|c|}{}                                    & 4500         & 6932.09 & 4380.19  \\ \hline \hline
\end{tabular}\vspace{1.2 mm}
\caption{Numerical values of final cumulative regrets of different algorithms under the experimental settings used in Figure~\ref{plt:exp} in Section~\ref{sec:exp} (strong adversaries).}\label{tab:exp}
\end{center}

\end{table}

\begin{table}[ht]
\begin{center}
\begin{tabular}{lcc|cc}
\hline \hline
\multicolumn{1}{l}{\multirow{10}{*}{Triangle reward function}} &  \multicolumn{1}{|c|}{Algorithm}                           & Budget ($C$) & Oracle   & Garcelon \\ \cline{2-5}
&\multicolumn{1}{|c|}{\multirow{3}{*}{Zooming}}            & 0            & 366.58   & 366.58   \\
&\multicolumn{1}{|c|}{}                                    & 3000         & 10861.72 & 10660.18 \\
&\multicolumn{1}{|c|}{}                                    & 4500         & 10862.75 & 10661.99 \\\cline{2-5}
&\multicolumn{1}{|c|}{\multirow{3}{*}{RMEL}}               & 0            & 512.46   & 512.46   \\
&\multicolumn{1}{|c|}{}                                    & 3000         & 624.29   & 620.96   \\
&\multicolumn{1}{|c|}{}                                    & 4500         & 623.50   & 634.59  \\\cline{2-5}
&\multicolumn{1}{|c|}{\multirow{3}{*}{BoB Robust Zooming}} & 0            & 461.16   & 461.16   \\
&\multicolumn{1}{|c|}{}                                    & 3000         & 545.27   & 561.77   \\
&\multicolumn{1}{|c|}{}                                    & 4500         & 552.66  & 569.51   \\ \hline \hline
\multicolumn{1}{l}{\multirow{10}{*}{Sine reward function}} & 
\multicolumn{1}{|c|}{Algorithm}                           & Budget ($C$) & Oracle  & Garcelon \\\cline{2-5}
&\multicolumn{1}{|c|}{\multirow{3}{*}{Zooming}}            & 0            & 315.94  & 315.94   \\
&\multicolumn{1}{|l|}{}                                    & 3000         & 5178.81 & 2636.73  \\
&\multicolumn{1}{|l|}{}                                    & 4500         & 5186.28 & 2799.22  \\ \cline{2-5}
&\multicolumn{1}{|c|}{\multirow{3}{*}{RMEL}}               & 0            & 289.86  & 289.86   \\
&\multicolumn{1}{|l|}{}                                    & 3000         & 280.62  & 277.22   \\
&\multicolumn{1}{|l|}{}                                    & 4500         & 284.94  & 288.06   \\ \cline{2-5}
&\multicolumn{1}{|c|}{\multirow{3}{*}{BoB Robust Zooming}} & 0            & 435.44  & 435.44   \\
&\multicolumn{1}{|c|}{}                                    & 3000         & 450.21  & 439.08   \\
&\multicolumn{1}{|c|}{}                                    & 4500         & 461.13 & 456.36  \\ \hline \hline
\multicolumn{1}{l}{\multirow{10}{*}{Two dim reward function}} & 
\multicolumn{1}{|c|}{Algorithm}                           & Budget ($C$) & Oracle  & Garcelon \\ \cline{2-5}
&\multicolumn{1}{|c|}{\multirow{3}{*}{Zooming}}            & 0            & 3248.54 & 3248.54  \\
&\multicolumn{1}{|l|}{}                                    & 3000         & 6380.37 & 6517.29  \\
&\multicolumn{1}{|l|}{}                                    & 4500         & 6991.41 & 6854.05 \\ \cline{2-5}
&\multicolumn{1}{|c|}{\multirow{3}{*}{RMEL}}               & 0            & 2589.32 & 2589.32  \\
&\multicolumn{1}{|l|}{}                                    & 3000         & 3198.06 & 2940.93  \\
&\multicolumn{1}{|l|}{}                                    & 4500         & 4231.88 & 4067.16  \\ \cline{2-5}
&\multicolumn{1}{|c|}{\multirow{3}{*}{BoB Robust Zooming}} & 0            & 3831.94 & 3831.94  \\
&\multicolumn{1}{|c|}{}                                    & 3000         & 4019.08 & 3335.67  \\
&\multicolumn{1}{|c|}{}                                    & 4500         & 4901.20 & 4054.05  \\ \hline \hline
\end{tabular}\vspace{1.2 mm}
\caption{Numerical values of final cumulative regrets of different algorithms under the experimental settings introduced in Appendix~\ref{app:exp} (weak adversaries).}\label{tab:exp2}
\end{center}

\end{table}